\documentclass[11pt]{article}
\usepackage[final]{acl}

\usepackage{times}
\usepackage{latexsym}
\usepackage[T1]{fontenc}
\usepackage[utf8]{inputenc}
\usepackage{microtype}
\usepackage{inconsolata}
\usepackage{graphicx}
\usepackage{float}
\usepackage{dblfloatfix}

\usepackage{url}
\usepackage{booktabs}
\usepackage{amsfonts}
\usepackage{amsmath}
\usepackage{nicefrac}
\usepackage{makecell}
\usepackage{tabularx}
\usepackage{multirow}
\usepackage[dvipsnames,svgnames,table]{xcolor}
\usepackage{tikz}
\usepackage{todonotes}
\usepackage{cleveref}
\usepackage{hyperref}
\usepackage{float}

\title{Do LLMs Choose Like Humans? \\ Using Cognitive Theory to Evaluate LLM Decision-Making}

\author{
Johnathan Sun$^{1,2}$ \qquad
Andrei Shleifer$^{1}$ \qquad
Yonatan Belinkov$^{2,3}$
\vspace{0.3em} \\
$^{1}$Harvard University \qquad
$^{2}$Kempner Institute, Harvard University \\
$^{3}$Technion - Israel Institute of Technology
\vspace{0.3em}\\
\texttt{jlsun@college.harvard.edu} \qquad
\texttt{shleifer@fas.harvard.edu} \\
\texttt{belinkov@technion.ac.il}
}

\begin{document}
\maketitle
\begin{abstract}
Large language models (LLMs) exhibit a range of human-like decision-making behaviors, but whether these reflect similar underlying mechanisms or surface-level mimicry remains unclear. We evaluate whether LLM context sensitivity aligns with a cognitive economic theory that explains human behavior through problem categorization and attention allocation. Across 12 open-source and commercial LLMs on a novel 140,000-trial product choice benchmark, context induces human-like shifts in choice and problem categorization, but does not reliably reweight attention between features like price and quality. Neither scale nor chain-of-thought reasoning reliably attenuates context sensitivity or generates human-like behavior. These results suggest that LLM decision mechanisms are distinct from human ones.
\end{abstract}

\section{Introduction}

Large language models (LLMs) increasingly make consequential economic decisions, but their underlying decision-making mechanisms remain poorly understood.

Behavioral evaluations have shown that models exhibit human-like preferences and cognitive biases \citep{malberg_comprehensive_2025, lior_comparing_2026, cook_what_2026}. Researchers have fit canonical economic models, such as prospect theory, on LLM behavior, but these theories offer no independent tests of mechanism and have so far shown limited explanatory power \citep{becker_irrational_1962, manski_identification_1995, ross_llm_2024, wang_rethinking_2026}.

Recent work in cognitive economics has made progress towards a unified explanation of human decision-making, including choice anomalies, cognitive biases, and preference instability \citep{bordalo_salience_2012, bordalo_salience_2013, bordalo_memory_2020}. We evaluate whether LLM decisions align with the cognitive theory of choice from \citet{bordalo_cognitive_2025}, and, by extension, the extent to which LLM decision-making is behaviorally and mechanistically consistent with humans.

We explore a central feature of decision-making: context sensitivity. Many behaviors, from canonical biases to phenomena traditionally modeled as belief biases or preference primitives, are instances of context sensitivity, so a sharp characterization of it in LLMs could help explain and predict behavior across many settings \citep{gabaix_behavioral_2019, jones_capturing_2022, enke_cognitive_2023}.

Is a jar of jam a luxury or a staple? Are stocks opportunities or gambles? \citet{bordalo_cognitive_2025} argue that humans make decisions by first characterizing the problem at hand: A decision-maker (DM) attends to context features like location and time to find the most similar category of choices made in the past. The selected category causes the DM to differentially attend to choice features, such as price and quality, which forms a mental representation of the current decision that is used to make a choice. Context influences how a person thinks about a problem, an explanation that unifies major biases in judgment and decision-making.

To evaluate whether LLMs exhibit behavior consistent with these mechanisms in a canonical economic setting, we introduce a 140,000-trial product choice dataset that experimentally varies context via cues that induce hedonic (quality-focused) or functional (price-focused) consumption in humans. We measure context effects on choices and feature sensitivities across 12 open-source and commercial LLMs, finding that:

\begin{enumerate}
    \item \textbf{Context effects on choice align with humans, but effects on feature sensitivities largely do not.} Cues have substantial effects on choice rates across all models, but only a quarter of model-cue combinations exhibit predicted shifts in feature sensitivity.
    \item \textbf{Context effects follow a problem categorization process.} Cue effects peak where categorization is ambiguous and shrink as baseline decisions grow more certain. Cues on prior in-context items predictably shift choices on a neutral test question.
    \item \textbf{Scale and reasoning do not reliably attenuate context effects.} Context sensitivity varies with scale across model families, remains high in frontier models, and can increase with chain-of-thought reasoning, contrasting with bounded-cognition accounts of human bias.
\end{enumerate}

\section{Related Work}

\paragraph{LLM decision-making and cognitive economics.}
LLMs exhibit a range of canonical biases and decision-making behaviors \citep{jones_capturing_2022, echterhoff_cognitive_2024, cheung_large_2025, malberg_comprehensive_2025}, and researchers have fit economic models on LLM behavior \citep{chen_emergence_2023, ross_llm_2024,bini_behavioral_2026}. We utilize a cognitive economic model that unifies context effects usually explained with domain-specific theories, which typically describe rather than mechanistically explain behavior, and often face documented experimental violations \citep{plott_willingness_2005, barr_case_2009, bernheim_behavioral_2018, dean_experimental_2023}. 

Cognitive economic models are potentially well-suited to LLMs. Such models are built from mechanisms of attention, memory, salience, and internal representations \citep{bordalo_salience_2012, bordalo_stereotypes_2016, bordalo_memory_2020, bordalo_cognitive_2025}, each with a natural analogue in transformers, which dynamically allocate attention across tokens \citep{vaswani_attention_2017}, use feed-forward layers as associative memory \citep{geva_transformer_2021}, and form context-sensitive hidden representations that shape decisions \citep{sun_persona_2026}.

\paragraph{Structure of context sensitivity.}
Models are highly sensitive to prompt content, including implicit demographic cues \citep{gupta_bias_2024, pawar_presumed_2025}, affective framing \citep{lior_comparing_2026}, and spurious syntax modifications \citep{sclar_quantifying_2024}. Our dataset is the first to measure the effects of hedonic and functional contexts, a longstanding distinction in the consumer choice literature \citep{dhar_consumer_2000}.

\paragraph{Similarity between LLMs and humans.}
Researchers in cognitive and social science increasingly use LLMs as experimental proxies for humans \citep{anthis_llm_2025, manning_general_2025, park_generative_2024}. However, human and LLM behavior often diverge in economic settings, and heuristic approaches such as prompt engineering or fine-tuning often fail to recover human-like behavior \citep{gui_challenge_2023, gao_take_2025, hullman_this_2026}. To the extent that \citet{bordalo_cognitive_2025} accurately describe humans, our results imply that underlying decision mechanisms in LLMs are distinct.

\section{Behavioral Framework}
\label{sec:theory}

\begin{figure*}[htbp!]
    \centering \includegraphics[width=\linewidth]{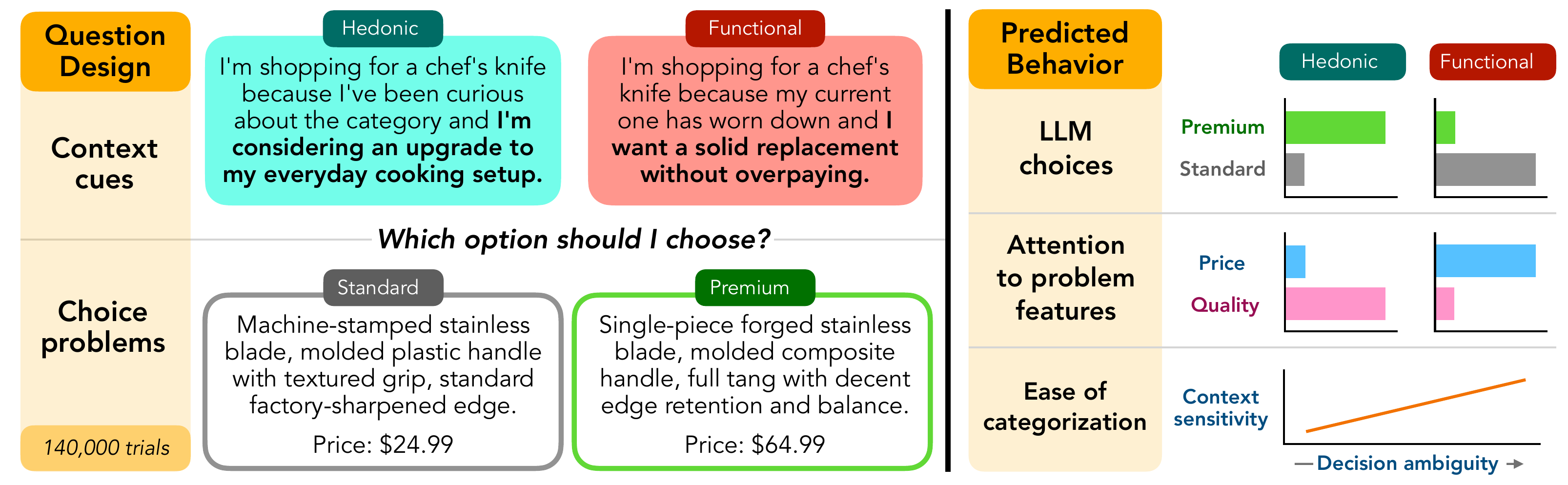}
    \caption{\textbf{Product choice questions evaluate LLM decisions against human behavior and cognitive theory.} (\textbf{Left}) Hedonic and functional context cues, drawn from the consumer choice literature, frame a binary choice between standard and premium products. (\textbf{Right}) Given human behavior explained by \citet{bordalo_cognitive_2025}, we predict that hedonic framing shifts LLM choices towards the premium option, reweights attention from price to quality, and produces larger effects when baseline categorization is ambiguous.}
    \label{fig:abstract}
\end{figure*}

\paragraph{Decision-makers weigh price and quality.} 

We evaluate LLMs on a fundamental economic setting: choosing between products. The tradeoff between price and quality is a canonical testbed in economics, with extensive human data and problem features that are easy to vary \citep{rosen_hedonic_1974, bordalo_salience_2013}. Moreover, LLMs increasingly encounter similar price-quality tradeoffs as they become full-fledged economic agents \citep{schumacher_agentic_2025, stripe_agentic_nodate}.

Our benchmark consists of product choice questions; we use the running example of purchasing jam to explain the mechanisms at play. Suppose a DM faces a problem $D$ in which they choose between a standard jar of jam $(j=0)$ and a fancy one $(j=1)$. Jam $j$ has price $p_j$ and quality $q_j$, with $p_1 > p_0$, $q_1 > q_0$. Let the DM's valuation of jam $j$ be
\begin{equation}
\label{eq:valuation}
u_j(\alpha^*) = \alpha_Q q_j - \alpha_P p_j,
\end{equation}
where $\alpha^*=(\alpha_Q,\alpha_P)$ are context-dependent attention weights on price and quality. The DM chooses the jam with the higher valuation, which is the fancy jam iff $\alpha_Q(q_1-q_0) > \alpha_P (p_1-p_0)$. 

\paragraph{Context affects price-quality tradeoff.}
\citet{bordalo_cognitive_2025} explain context-sensitive behavior in humans by specifying how attention weights $(\alpha_Q, \alpha_P)$ are pinned down in the mental representation of a decision.

When is jam a luxury or a staple? Consumer psychology models propose two ways of looking at the problem \citep{hirschman_hedonic_1982, dhar_consumer_2000}:
\begin{enumerate}
    \item \textbf{Hedonic consumption} implicates affective or sensory experiences associated with quality sensitivity: Jam as a luxury.
    \item \textbf{Functional consumption} implicates accomplishing a practical task associated with price sensitivity: Jam as a staple.
\end{enumerate}
The core intuition is that context influences whether the DM views a decision as hedonic or functional consumption, which changes how the DM weighs price and quality. 

For example, the shopping venue influences whether humans characterize their purchasing as functional or hedonic \citep{hamilton_low_2013}. To capture this effect, let $\kappa_D$ denote the scalar context of the current problem $D$. Suppose $\kappa_D = 1$ in an organic farm store (hedonic), and $\kappa_D = 0$ in a convenience store (functional). Let $F_{\text{hed}}$ and $F_{\text{fun}}$ denote the frequencies of hedonic and functional consumption in past problems, which represent experiences stored in memory. The DM chooses a jam to purchase in two steps:
\begin{enumerate}
    \item \textbf{Problem categorization}: The DM selects a category from memory that is most contextually similar to the current problem. Here, the DM categorizes $D$ as hedonic if $\kappa_D > \kappa(F_{\text{fun}} / F_{\text{hed}})$ and functional otherwise, where $\kappa(\cdot)$ is increasing. Intuitively, the DM is more likely to regard the jam decision as hedonic consumption in the farm store because they were hedonic there in the past.
    \item \textbf{Attention reweighting}: The DM forms a mental representation with the selected category, which influences relative attention to decision features. If $D$ is hedonic, $\alpha_Q > \alpha_P$, so the DM is more sensitive to quality than price. If $D$ is functional, $\alpha_P > \alpha_Q$.
\end{enumerate}

\paragraph{Model implications.}
Hedonic/functional categorization predicts attribute weighting, price elasticity, and purchase justification in humans \citep{wakefield_situational_2003, okada_justification_2005, khan_licensing_2006}. We evaluate LLMs against three human phenomena explained by the cognitive model:

\begin{enumerate}
    \item \textbf{Context affects choices}: Under \Cref{eq:valuation}, hedonic context raises $\alpha_Q$ relative to $\alpha_P$, which shifts choices near indifference towards the premium option. The DM is more likely to choose the fancy jam in the farm store, and vice versa in the convenience store.

    \item \textbf{Context induces asymmetric feature sensitivity}:
    Hedonic context raises the marginal rate of substitution $\alpha_Q / \alpha_P$, the amount of additional price the DM accepts for a marginal gain in quality. Intuitively, the DM's decision is relatively more sensitive to jam qualities than jam prices in the farm store.

    \item \textbf{Context effects depend on ease of categorization}: Holding context fixed, the DM is more likely to categorize $D$ as hedonic if $F_{\text{fun}} / F_{\text{hed}}$ is smaller, which occurs when the hedonic category has been used relatively more often than the functional category. The DM is more likely to view jam as a luxury if they made relatively more recent hedonic purchases.
\end{enumerate}

\section{Experimental Design}
\label{sec:exp_design}

\begin{table*}[ht!]
\centering
\footnotesize
\renewcommand{\arraystretch}{1.2}
\begin{tabular}{p{2.3cm} p{4.7cm} p{4.7cm} p{2.5cm}}
\toprule
\textbf{Cue Type} & \textbf{Hedonic Cue} & \textbf{Functional Cue} & \textbf{References} \\
\midrule
Purchase trigger & Exploring the category out of curiosity or considering an upgrade & Routine replacement or replenishment without overpaying & \citet{bell_point_2011} \\
Retail architecture & Browsing a curated specialty page with staff picks and quality notes & Browsing a marketplace comparison page sorted by price with filters & \citet{koschmann_retailer_2018} \\
Payment source & Paying with a gift card or credit, keeping any unspent balance & Paying from a checking account, keeping any unspent balance & \citet{thaler_mental_1999} \\
Justification & Recently earned a small reward; this feels like a reasonable treat & Ordinary purchase with no special reason; just a practical choice & \citet{okada_justification_2005} \\
Reasoning frame & Focused on why the product would improve the user experience & Focused on how to complete the purchase and the concrete tradeoffs & \citet{liberman_role_1998} \\
\bottomrule
\end{tabular}
\caption{\textbf{Contextual purchasing cues.} Each cue type includes a hedonic/functional pair drawn from the consumer choice literature. Claude Sonnet-4.6 instantiates these templates into concrete cues for each product, designed to shift consumption framing without changing product utility.}
\label{tab:cues}
\end{table*}

\paragraph{Baseline product choice problems.}
To evaluate LLM behavior, we design a 140,000-trial product choice dataset. We use Claude Sonnet-4.6 to generate 100 products (``chef's knife'') from 20 sectors (``kitchen \& food''). For each product, Claude generates five options that gradually increase in price $p_t$ and descriptive quality $d_t$. Descriptions match in length ($\pm 5$ words) and attributes described per product, and brand names are omitted to limit associations with unobservable features like reputation. 

To map description to quality, we use GPT 5.4 to generate 0-100 quality scores $q_t$ for each option. GPT is employed so Claude does not self-evaluate its quality manipulations, though GPT and Claude scores are highly correlated ($r = 0.96$). We manually audited 10\% of GPT scores to confirm that better-described options consistently received higher scores, and that score gaps tracked the size of described quality differences.

As in the jam example, we write questions where an LLM with context $\kappa$ outputs a binary choice A/B between standard $(j=0)$ and premium $(j=1)$ options for a given product, with $p_1 > p_0$, $q_1 > q_0$. For each product, we pair the five options by adjacent quality, and include both option orders to account for position effects. This process yields 800 total pairs of product options.

\paragraph{Context cues drawn from hedonic/functional consumption literature.}
Context $\kappa$ varies under three conditions: The \textbf{neutral} condition presents the statement ``I need a \{product\}'' before the question. The \textbf{hedonic} and \textbf{functional} conditions prepend one of five types of paired cues to the neutral statement. Claude Sonnet-4.6 generates cues using templates that induce or are associated with hedonic/functional consumption in the consumer choice literature, such as whether a DM intends to replace (functional) or upgrade (hedonic) their existing product (\Cref{tab:cues}). With five types of paired hedonic/functional cues and one neutral condition, the baseline dataset contains 8,800 problems. \Cref{fig:abstract} presents an example problem, and Appendix \Cref{app:design} includes more technical dataset details.

\paragraph{Datasets to estimate feature sensitivity.}
We estimate the sensitivity of choices to problem features using two additional datasets that slightly perturb the price and quality of baseline questions. The \textbf{price sweep} dataset fixes descriptions and scales premium option prices by 10 evenly log-spaced factors from one to two, generating 88,000 problems. The \textbf{quality sweep} dataset fixes prices and varies premium option descriptions by five small differences in quality, adding a further 44,000 problems.

\begin{figure*}[t]
    \centering \includegraphics[width=\linewidth]{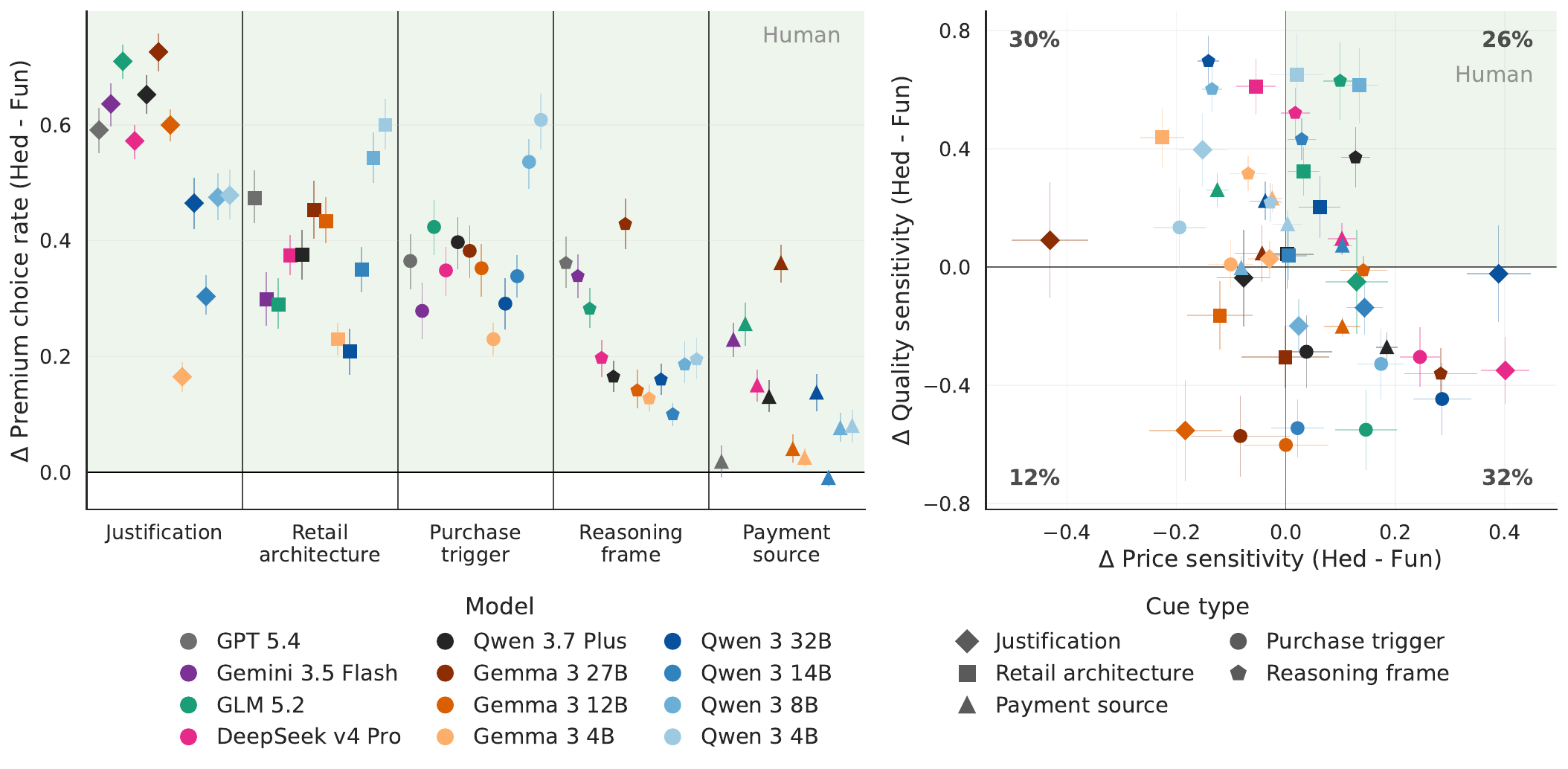}
    \caption{\textbf{Models match human context effects on choices, but diverge on feature sensitivity.} (\textbf{Left}) Difference in premium choice rates between hedonic and functional cues. Shaded region indicates human-like behavior for all panels---here, hedonic cues lead to more premium choices. See Appendix \Cref{fig:appendix_hed_fun} for the difference in premium choice rates relative to the neutral condition. (\textbf{Right}) Difference in standardized price (x-axis) and quality (y-axis) sensitivity between hedonic and functional cues, estimated by linear regression on choice logits (see \Cref{sec:exp_design}). Shaded region denotes behavior where hedonic cues attenuate (negative) price sensitivity and amplify quality sensitivity. Percentages report the share of model-cue combinations in each quadrant. All bars are 95\% CIs.}
    \label{fig:choice_rate}
\end{figure*}

To investigate how cues shift feature sensitivity, we measure decisions as a function of log feature ratios, a form that maps the marginal rate of substitution $\alpha_Q/\alpha_P$ onto a regression coefficient. For each model $m$ and sweep $s \in \{\text{price}, \text{quality}\}$, we estimate the regression
\begin{align*}
\ell_i &= \beta_H\mathrm{Hed}_i + \beta_R \log
  r_i^{(s)} + \beta_{HR}\mathrm{Hed}_i \log
  r_i^{(s)} \\
  &+ \alpha_{p(i)} + \gamma \mathrm{PF}_i + \varepsilon_i,
\end{align*}
where $\ell_i$ is the logit difference between premium and standard options; $\mathrm{Hed}_i \in \{0, 1\}$ indicates a hedonic cue (functional is the reference); $\log r_i^{(\text{price})}$ is the log-multiplier applied to the premium price in trial $i$, while $\log r_i^{(\text{quality})} = \log(q^{\text{prem}}_i / q^{\text{std}}_i)$ is the log ratio of premium-to-standard quality scores; $\alpha_{p(i)}$ are option pair and cue type fixed effects; and $\mathrm{PF}_i$ controls for option order. The coefficient $\beta_R$ captures the model's sensitivity to feature ratios under functional cues, and $\beta_H$ captures the average level shift induced by hedonic cues. The key coefficient is the interaction $\beta_{HR}$: the marginal change in feature sensitivity between hedonic and functional cues. Human evidence and theory predict $\beta_{HR} > 0$ in both price and quality sweeps, which corresponds to hedonic cues attenuating negative price sensitivity and amplifying positive quality sensitivity relative to functional cues.

\paragraph{Models.}
We evaluate 12 instruct-tuned open-source and commercial LLMs: GPT 5.4, Gemini 3.5 Flash, GLM 5.2, DeepSeek v4 Pro, Qwen 3.7 Plus, Gemma 3 (4B, 12B, and 27B), and Qwen 3 (4B, 8B, 14B, and 32B) \citep{google_deepmind_gemini_2025, team_gemma_2025, yang_qwen3_2025, openai_gpt_2026, glm-5-team_glm-5_2026, deepseek-ai_deepseek-v4_2026}. For the main experiments, we query each model once per question at temperature zero. GPT and Gemini, for which temperature controls and logits are unavailable, are run with medium (default) reasoning and excluded from the feature-sensitivity analysis in \Cref{sec:sensitivity}. In \Cref{sec:scaling}, we additionally query the four Qwen 3 models once per question in thinking mode with a 4096-token budget at temperature 0.6.

\section{Comparing LLM and Human Behavior}

\subsection{Human-LLM similarity in choice rates}

Every LLM is more likely to choose premium with hedonic cues than functional cues, consistent with context effects on human choice (\Cref{fig:choice_rate}, left). This effect is significant ($p < 0.001$, one-sided paired $t$-test), robust across almost every model-cue combination, and substantial: The average LLM is more than twice as likely to choose premium with a hedonic cue as with a functional cue.

Effect sizes vary across cue types. The hedonic justification cue---buying a special treat instead of a normal purchase---raises premium choice rates the most relative to the corresponding functional cue (248\% increase). The hedonic payment source cue---paying with a gift card instead of a checking account---has the smallest effect (36\% increase), though similar interventions induce substantial changes in human behavior \citep{levav_emotional_2009, helion_gift_2014}. The moderate effect of hedonic reasoning frame cues is also unexpected, since these cues effectively instruct the LLM to adopt a consumption mode.

Finally, hedonic (functional) cues raise (lower) premium choice rates relative to the neutral condition for every LLM ($p < 0.05$, one-sided paired $t$-test), so both cue directions significantly impact choices. That said, the neutral premium choice rate is 2.8 times closer on average to the hedonic-cue rate than to the functional-cue rate. Hence, the hedonic-functional gap in premium choice is largely because functional cues alter premium-leaning ``default'' behavior.

\subsection{Human-LLM contrast in feature sensitivity}
\label{sec:sensitivity}

\paragraph{Sensitivities diverge from human-like behavior.}
While LLMs demonstrate human-like context effects on choices, they do not reweight feature sensitivities similarly. Only 26\% of model-cue combinations are less price sensitive and more quality sensitive in hedonic cues compared to functional, while 12\% exhibit the opposite behavior in both (\Cref{fig:choice_rate}, right). The remaining 62\% of model-cue combinations exhibit human-like reweighting in one feature, but not the other. These distinct effects on feature sensitivity are robust to different regression specifications, such as estimating sensitivity to differences in feature values instead of ratios (Appendix \Cref{fig:appendix_regression_robustness}).

\paragraph{Context effects on feature sensitivity are highly heterogeneous.}
Effects vary significantly by cue type. For instance, Qwen 32B is more price and quality sensitive under hedonic reasoning frame cues, but less sensitive to both under hedonic purchase trigger cues. No LLM exhibits human-like sensitivity across all cue types. 

Effects also vary by model: hedonic reasoning frame cues reduce both price and quality sensitivity in Gemma 27B, but raise both in Qwen 8B. No cue type induces human-like sensitivity effects in all LLMs, though the retail architecture cues are most successful (6/10 models) and the purchase trigger cues least (0/10 models). Effects on feature sensitivity span at least three quadrants of \Cref{fig:choice_rate} (right) for six out of ten LLMs.

\paragraph{Aggregating models does not recover human-like sensitivities.}
Context effects may appear stable across a human population but vary between individuals, domains, and problems \citep{liew_appropriacy_2016}. Similarly, while individual LLM behavior is noisy, population-level behavior may be more human-like \citep{hagendorff_human-like_2023}. However, aggregating LLMs for each cue type reveals that only reasoning frame cues induce significant human-like sensitivities, while only four out of ten models display this pattern when aggregating across cues (GLM 5.2; DeepSeek v4 Pro; Qwen 8B and 32B).

\paragraph{Despite diverging sensitivities, LLMs do choose coherently.}
Another concern is that sensitivity patterns reflect incoherent choices rather than meaningful differences from human behavior. However, feature sensitivities are tightly estimated, and LLM choices have meaningful structure. Every LLM exhibits negative price sensitivity and positive quality sensitivity under both hedonic and functional cues. Holding quality fixed, doubling the premium price ratio reduces premium choice by 15 pp.\ on average; holding price fixed, doubling the quality ratio raises it by 44 pp. While changes to feature sensitivities from context vary, LLMs do respond consistently to feature variation.

\subsection{LLMs are sensitive to ease of categorization}
\label{sec:categorization}

\begin{figure*}[!t]
    \centering \includegraphics[width=\linewidth]{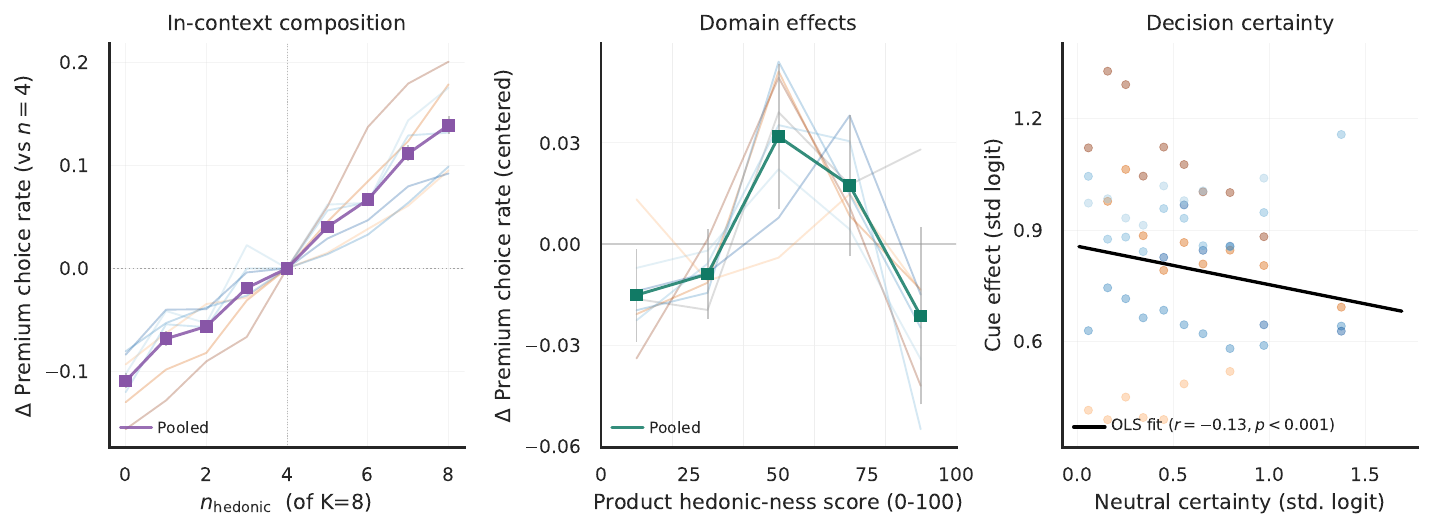}
    \caption{\textbf{Context effects follow a categorization process: in-context evidence shifts categorization (left), and ambiguity amplifies cue effects (center, right).} Panels pool across Gemma 3 (4B, 12B, 27B) and Qwen 3 (4B, 8B, 14B, 32B). (\textbf{Left}) Difference in premium choice rate relative to $n_{\text{hed}} = 4$ (midpoint) as a function of the number of hedonic cues among eight in-context examples. (\textbf{Center}) Difference in premium choice rate between hedonic and functional cues (de-meaned by model) binned by product hedonic-ness scores. (\textbf{Right}) Difference in (premium - standard logit) between hedonic and functional cues as a function of (premium - standard logit) in neutral choice problems, both standardized to neutral-SD units. All bars are 95\% CIs.}
    \label{fig:frequency}
\end{figure*}

Since LLMs do not exhibit human-like feature reweighting, we ask whether LLMs behave consistently with the preceding mechanism: problem categorization. Recall from \Cref{sec:theory} that a DM categorizes problem $D$ as hedonic if $\kappa_D > \kappa(F_{\text{fun}}/F_{\text{hed}})$. If LLMs categorize problems, we expect three behaviors: (i) Directly manipulating $F_{\text{hed}}/F_{\text{fun}}$ in-context shifts choices. (ii) Context effects on choice attenuate in problems with obvious categorizations. (iii) Context effects on choice are largest when decisions are ambiguous, irrespective of category taxonomy. We find evidence of all three patterns across models.

\paragraph{In-context evidence shifts categorization.}
Raising $F_{\text{hed}}/F_{\text{fun}}$ lowers the threshold for categorizing a current problem as hedonic, which captures the intuition that humans are generally more likely to recall and act upon frequent and recent memories \citep{kahana_foundations_2012, bordalo_ads_2025, jiang_investor_2025}. To proxy memory, we manipulate $F_{\text{hed}}/F_{\text{fun}}$ in LLMs by prepending eight cued example problems before an un-cued test problem. The eight examples include $n_{\text{hed}} \in \{0, 1, \ldots, 8\}$ hedonic and $n_{\text{fun}} = 8 - n_{\text{hed}}$ functional cues of the same type, present different products from the test question, and omit choices to avoid pattern-matching decisions. Consistent with the theory, increasing $n_{\text{hed}}$ from zero to eight raises the average premium choice rate from 49\% to 74\%, and a 10 pp.\ increase in the share of hedonic examples raises premium choice rates by 2.5 pp.\ on average (\Cref{fig:frequency}, left).

\paragraph{Implicit associations moderate context effects.} 
In humans, buying a fountain pen feels more hedonic than buying dish soap \citep{wakefield_situational_2003}. Since LLMs train on vast amounts of human data, products themselves may be associated with types of consumption, generating a test of categorization \citep{caliskan_semantics_2017, li_frontiers_2024}.

We can decompose context as $\kappa_D(p, c) = x_p + \gamma_c$, where $x_p$ is context conveyed by product $p$ and $\gamma_c$ is context from an explicit cue $c$. The cue can only change categorization if $\kappa_D$ is sufficiently close to the threshold $\tau \in [x_p + \gamma_{\text{fun}}, x_p + \gamma_{\text{hed}}]$. For strongly hedonic or functional products, both cues produce the same categorization, attenuating their effect on choice.

To test this behavioral signature, we use a GPT 5.4-based judge to rate each product's name from 0 (fully functional) to 100 (fully hedonic), averaging over three trials \citep{asirvatham_gpt_2026}. We manually audited all scores, and Appendix \Cref{fig:appendix_hedonic_scores} includes the score distribution. Measuring tape and dish soap are the most functional products, while watercolor paints and greeting cards are the most hedonic. Behavior is consistent with our prediction: cue effects on choices across LLMs peak for products with midrange hedonic-ness ratings and shrink at both extremes (\Cref{fig:frequency}, center).

\paragraph{Decision ambiguity amplifies context effects.}
The hedonic-functional analysis assumes that these are the two relevant problem categories. A weaker test of categorization is that context should matter most when an un-cued problem does not induce a stable representation and choice. We proxy categorization ambiguity with the decision certainty of LLMs under neutral cues, measured as the absolute logit difference between premium and standard options (larger = more certain). Across models, cue effects on choice rates are negatively associated with baseline certainty (\Cref{fig:frequency}, right). 

While our interpretation assumes that competing categories imply different choices (e.g., hedonic and premium, functional and standard), this behavioral pattern is hard to reconcile with cues generating fixed shifts in the price-quality tradeoff. We propose that when an LLM is uncertain about how to categorize a problem, a cue provides a greater update to its posterior beliefs. Similar logic motivates Bayesian and rational inattention accounts of context dependence, and is consistent with evidence in humans that comparison complexity and cognitive uncertainty drive context dependence and preference instability \citep{enke_cognitive_2023, shubatt_tradeoffs_2026}.

\begin{figure*}[!t]
    \centering \includegraphics[width=\linewidth]{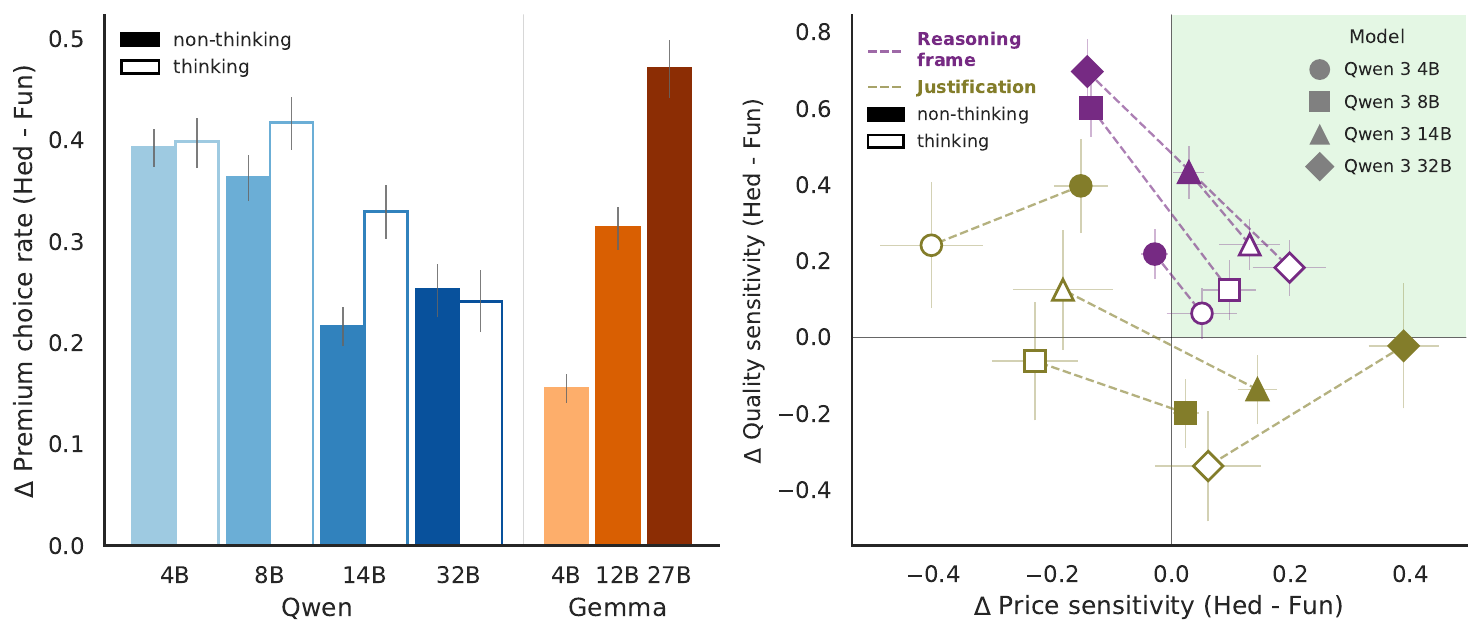}
    \caption{\textbf{Scale and reasoning do not reliably reduce context sensitivity.} (\textbf{Left}) Difference in premium choice rate between hedonic and functional cues for Qwen 3 (paired non-thinking/thinking) and Gemma 3 (non-thinking) model families. (\textbf{Right}) Difference in standardized price (x-axis) and quality (y-axis) sensitivity between hedonic and functional cues for Qwen models in thinking/non-thinking modes. Shaded region denotes human-consistent behavior. All bars are 95\% CIs.}
    \label{fig:scaling}
\end{figure*}

\subsection{Scale and reasoning do not consistently reduce context sensitivity}
\label{sec:scaling}

\citet{bordalo_cognitive_2025} explain human context sensitivity in part through limited attention, and a broader literature links such behaviors to bounded cognition and computational constraints \citep{stanovich_individual_2000, gabaix_behavioral_2019, mackowiak_rational_2023}. If the same constraint shapes LLM decisions, more capable models should be less context-sensitive. However, two margins of capability---scale and inference-time compute---do not consistently attenuate context effects on choices. Reasoning can recover human-like context effects on feature sensitivities, but only when cues explicitly invoke the hedonic/functional dimension.

\paragraph{Context effects on choice vary with scale across LLM families.}
We evaluate context effects on choice for four Qwen 3 models and three Gemma 3 models. The two largest Qwen models are less sensitive than the two smaller ones, but Gemma~27B is roughly three times more sensitive than Gemma~4B (\Cref{fig:scaling}, left). Context effects for GPT and Gemini, two commercial frontier LLMs, are also larger than those of Qwen~32B on average. One potential explanation is that post-training pressure---instruction tuning, RLHF, or training to infer user attributes---may work against any rationality gained with scale \citep{itzhak_instructed_2024}.

\paragraph{Reasoning recovers human-like behavior only when cues are explicit.}
We evaluate the four Qwen 3 models in thinking mode. Chain-of-thought reasoning does not reduce context effects on choice and significantly increases them for Qwen~8B and~14B (\Cref{fig:scaling}, left).

The effect of reasoning on feature sensitivity varies by cue. The reasoning frame cue, which effectively instructs an LLM to engage in hedonic or functional consumption, induces human-like sensitivity effects in all four Qwen models only in thinking mode (\Cref{fig:scaling}, right). The effect of reasoning is very inconsistent for other cues; we provide the justification cue as an example in \Cref{fig:scaling} and include all five cue types in Appendix \Cref{fig:appendix_qwen_impact}. Human-like sensitivity effects may be achievable, but they appear to be fragile, consistent with previous research on human simulation with LLMs \citep{hullman_this_2026}.

\section{Discussion}

\paragraph{Decision mechanisms in LLMs are distinct from those in humans.}
While context effects on choice and problem categorization are consistent with the theory of \citet{bordalo_cognitive_2025} and human behavior, feature sensitivities are not. Notably, LLMs do not exhibit human-like instability in price elasticities \citep{wakefield_situational_2003}. 

Our results contribute to a growing literature on the potential hazards of treating LLMs as human proxies \citep{gui_challenge_2023, gao_take_2025}. In particular, researchers should be cautious when using comparative statics from LLM choice data to evaluate claims about humans. Agreement between human and LLM choices, when used alone, does not identify mechanistic correspondence.

That said, LLMs have made progress in accurately predicting human behavior, whether by default or with deliberate scaffolding, across social and cognitive science \citep{binz_foundation_2025, ashokkumar_large_2026}. An important avenue of research is understanding why LLMs seem to behave like humans in some settings, while sharply diverging from humans in others.

For instance, \citet{bordalo_cognitive_2025} argue that categories arise because humans selectively recall memories from past experiences---including context, reasoning, and decisions---when encountering a new problem. We lack a rigorous understanding of whether LLMs encode analogous ``experiences'' about choice from training, and whether they use similar recall mechanisms in decision-making. To that end, training LLMs with targeted human data or structuring LLM representations to mimic those of humans could be promising steps towards faithful simulation \citep{binz_foundation_2025}.

\paragraph{Understanding LLM decision-making on its own terms.}
As LLMs become full-fledged economic agents, their decision representations, preferences, and mechanisms are becoming highly relevant objects of study. Our work is an example of using economic reasoning to understand LLM behavior \citep{imas_agentic_2025, cook_what_2026, buchanan_innate_2026}. While LLMs may not behave fully like humans, we show that their decisions are consistent with utility maximization and problem categorization and that context predictably shifts their choices.

Our understanding of LLM behavior is much less developed than for humans, which opens several avenues for future work. First, the magnitude and direction of context effects, especially sensitivity to problem features, vary substantially across tested cues and models. We hope to study this heterogeneity in future work, including in more naturalistic settings and other types of decision problems.

Second, unlike human studies, we can directly measure and intervene on the internal representations of choice problems across contexts. Interpretability methods could complement behavioral studies by causally identifying circuits responsible for context sensitivity and bias, which may help disambiguate between competing theories and develop de-biasing interventions \citep{vig_causal_2020, geiger_finding_2024, karvonen_robustly_2025}.

Finally, our experiments are grounded in the bias literature. While existing human studies provide a search heuristic for LLM failures, biases and behaviors \textit{unique} to LLMs remain under-explored \citep{jones_capturing_2022, stella_using_2023, mccoy_embers_2023, cheung_large_2025}. One potential explanation for our results is implicit personalization: LLMs make decisions given user beliefs and preferences inferred from context, which may explain why the frontier models in our sample are the most sensitive to context \citep{xie_explanation_2022, marks_persona_2026}.

\section*{Limitations}

\paragraph{Behavioral similarity does not imply mechanistic correspondence.}
We evaluate LLM behavior against empirical human regularities and the behavioral predictions of a cognitive theory. While divergences between LLM and human behavior do imply distinct underlying mechanisms, similarities in behavior, such as context effects on choices or evidence on problem categorization in \Cref{sec:categorization}, do not imply shared mechanisms. Other theories may generate similar predictions about behavior; we do not disambiguate between competing explanations in this work. Mechanistic evidence from probing, steering, or causal interventions on internal representations is likely necessary to identify the processes that drive the behaviors we observe. 

\paragraph{Experiments are imperfect analogies between humans and LLMs.}
Attention in humans has been measured through physical methods such as eye tracking or through self-reported scores \citep{reutskaja_search_2011, bordalo_how_2025}. The human-like interpretation of token-level attention is tenuous, so we estimate LLM attention via regression as implied weights on problem features, a standard technique in economics \citep{bibal_is_2022}. In our experiments, higher attention means that an LLM behaves as if a feature is more important for choice. We leave the proper measurement and interpretation of LLM attention for future work. Likewise, our experiments in \Cref{sec:categorization} treat in-context examples as loosely analogous to human experiences in memory. In-context examples are not the same as training data, and both may fundamentally differ from human memories.

A related limitation is that our predictions are taken from cognitive theory and existing human studies rather than a matched LLM-human experiment with identical questions. Although we use cues closely borrowed from the consumer choice literature, our cues may differ in systematic ways from those in prior work.

\paragraph{Limited scope of comparison.}
We test one canonical economic setting: binary choice between standard and premium consumer products. The theory of \citet{bordalo_cognitive_2025} applies to a much broader class of decision problems, including risk, intertemporal choice, and social preferences. We encourage future work in these domains, and with more LLMs and theories, to investigate the underlying mechanisms of LLM decision-making.

\section*{Supplemental Materials}

This work’s supplemental materials, including questions, model responses, and analysis code, are available at \href{https://github.com/johnathansun/llm-cognitive-choice}{github.com/johnathansun/llm-cognitive-choice}.

\section*{Acknowledgments}

We thank Joshua Schwartzstein, Pedro Bordalo, Suproteem Sarkar, and Victoria Li for helpful comments and discussions. This work has been made possible in part by a gift from the Chan Zuckerberg Initiative Foundation to establish the Kempner Institute for the Study of Natural and Artificial Intelligence at Harvard University. In addition, it was supported by the Israel Science Foundation (grant No.\ 2942/25) and the European Union (ERC, Control-LM, 101165402). Views and opinions expressed are however those of the author(s) only and do not necessarily reflect those of the European Union or the European Research Council Executive Agency. Neither the European Union nor the granting authority can be held responsible for them. 

\bibliography{references}

@inproceedings{anthis_llm_2025,
  title     = {Position: {LLM} Social Simulations Are a Promising Research Method},
  author    = {Anthis, Jacy Reese and Liu, Ryan and Richardson, Sean M. and Kozlowski, Austin C. and Koch, Bernard and Brynjolfsson, Erik and Evans, James and Bernstein, Michael S.},
  booktitle = {Proceedings of the 42nd International Conference on Machine Learning},
  editor    = {Singh, Aarti and Fazel, Maryam and Hsu, Daniel and Lacoste-Julien, Simon and Berkenkamp, Felix and Maharaj, Tegan and Wagstaff, Kiri and Zhu, Jerry},
  series    = {Proceedings of Machine Learning Research},
  volume    = {267},
  pages     = {81005--81034},
  year      = {2025},
  publisher = {PMLR},
  url       = {https://proceedings.mlr.press/v267/anthis25a.html}
}

@article{ashokkumar_large_2026,
    title = {Large language models can predict the results of social science experiments},
    author = {Ashokkumar, Ashwini and Hewitt, Luke and Ghezae, Isaias and Willer, Robb},
    journal = {Nature},
    volume = {656},
    number = {8126},
    pages = {115--122},
    month = aug,
    year = {2026},
    doi = {10.1038/s41586-026-10742-x},
    url = {https://www.nature.com/articles/s41586-026-10742-x},
}

@techreport{asirvatham_gpt_2026,
    title = {{GPT} as a measurement tool},
    author = {Asirvatham, Hemanth and Mokski, Elliott and Shleifer, Andrei},
    type = {Working Paper},
    number = {34834},
    institution = {National Bureau of Economic Research},
    month = feb,
    year = {2026},
    doi = {10.3386/w34834},
    url = {https://www.nber.org/papers/w34834},
}

@incollection{barr_case_2009,
    title = {The case for behaviorally informed regulation},
    booktitle = {New {Perspectives} on {Regulation}},
    editor = {Moss, David and Cisternino, John},
    publisher = {The Tobin Project},
    address = {Cambridge, MA},
    author = {Barr, Michael S. and Mullainathan, Sendhil and Shafir, Eldar},
    year = {2009},
    chapter = {2},
    pages = {25--61},
    url = {https://tobinproject.org/sites/default/files/assets/New_Perspectives_Ch2_Barr_Mullainathan_Shafir.pdf},
}

@article{becker_irrational_1962,
    title = {Irrational behavior and economic theory},
    volume = {70},
    number = {1},
    journal = {Journal of Political Economy},
    author = {Becker, Gary S.},
    month = feb,
    year = {1962},
    pages = {1--13},
    doi = {10.1086/258584},
}

@article{bell_point_2011,
    title = {From point of purchase to path to purchase: {How} preshopping factors drive unplanned buying},
    volume = {75},
    number = {1},
    doi = {10.1509/jm.75.1.31},
    journal = {Journal of Marketing},
    author = {Bell, David R. and Corsten, Daniel and Knox, George},
    month = jan,
    year = {2011},
    pages = {31--45},
}

@incollection{bernheim_behavioral_2018,
    title = {Behavioral public economics},
    booktitle = {Handbook of {Behavioral} {Economics}: {Applications} and {Foundations} 1},
    volume = {1},
    editor = {Bernheim, B. Douglas and DellaVigna, Stefano and Laibson, David},
    publisher = {North-Holland},
    author = {Bernheim, B. Douglas and Taubinsky, Dmitry},
    year = {2018},
    pages = {381--516},
    doi = {10.1016/bs.hesbe.2018.07.002},
    url = {https://www.sciencedirect.com/science/article/pii/S2352239918300022},
}

@inproceedings{bibal_is_2022,
    title     = {Is Attention Explanation? {An} Introduction to the Debate},
    author    = {Bibal, Adrien and Cardon, R{\'e}mi and Alfter, David and Wilkens, Rodrigo and Wang, Xiaoou and Fran{\c{c}}ois, Thomas and Watrin, Patrick},
    editor    = {Muresan, Smaranda and Nakov, Preslav and Villavicencio, Aline},
    booktitle = {Proceedings of the 60th Annual Meeting of the Association for Computational Linguistics (Volume 1: Long Papers)},
    month     = may,
    year      = {2022},
    address   = {Dublin, Ireland},
    publisher = {Association for Computational Linguistics},
    pages     = {3889--3900},
    doi       = {10.18653/v1/2022.acl-long.269},
    url       = {https://aclanthology.org/2022.acl-long.269/}
}

@techreport{bini_behavioral_2026,
    title = {Behavioral economics of {AI}: {LLM} biases and corrections},
    author = {Bini, Pietro and Cong, Lin William and Huang, Xing and Jin, Lawrence J.},
    type = {Working Paper},
    number = {34745},
    institution = {National Bureau of Economic Research},
    month = jan,
    year = {2026},
    doi = {10.3386/w34745},
    url = {https://www.nber.org/papers/w34745},
}

@article{binz_foundation_2025,
    title = {A foundation model to predict and capture human cognition},
    author = {Binz, Marcel and Akata, Elif and Bethge, Matthias and Brändle, Franziska and Callaway, Fred and Coda-Forno, Julian and Dayan, Peter and Demircan, Can and Eckstein, Maria K. and Éltető, Noémi and Griffiths, Thomas L. and Haridi, Susanne and Jagadish, Akshay K. and Ji-An, Li and Kipnis, Alexander and Kumar, Sreejan and Ludwig, Tobias and Mathony, Marvin and Mattar, Marcelo and Modirshanechi, Alireza and Nath, Surabhi S. and Peterson, Joshua C. and Rmus, Milena and Russek, Evan M. and Saanum, Tankred and Schubert, Johannes A. and Schulze Buschoff, Luca M. and Singhi, Nishad and Sui, Xin and Thalmann, Mirko and Theis, Fabian J. and Truong, Vuong and Udandarao, Vishaal and Voudouris, Konstantinos and Wilson, Robert and Witte, Kristin and Wu, Shuchen and Wulff, Dirk U. and Xiong, Huadong and Schulz, Eric},
    journal = {Nature},
    volume = {644},
    number = {8078},
    pages = {1002--1009},
    month = aug,
    year = {2025},
    doi = {10.1038/s41586-025-09215-4},
    url = {https://www.nature.com/articles/s41586-025-09215-4},
}

@techreport{bordalo_ads_2025,
    title = {Ads as cues},
    type = {Working Paper},
    number = {34387},
    institution = {National Bureau of Economic Research},
    url = {https://www.nber.org/papers/w34387},
    doi = {10.3386/w34387},
    author = {Bordalo, Pedro and Burro, Giovanni and Gennaioli, Nicola and Nacamulli, Gad and Shleifer, Andrei},
    month = oct,
    year = {2025},
}

@article{bordalo_cognitive_2025,
    title = {A cognitive theory of reasoning and choice},
    volume = {141},
    number = {3},
    url = {https://doi.org/10.1093/qje/qjag023},
    doi = {10.1093/qje/qjag023},
    journal = {The Quarterly Journal of Economics},
    author = {Bordalo, Pedro and Gennaioli, Nicola and Lanzani, Giacomo and Shleifer, Andrei},
    month = aug,
    year = {2026},
    pages = {1921--1963},
}

@article{bordalo_how_2025,
    title = {How people use statistics},
    volume = {93},
    number = {1},
    url = {https://doi.org/10.1093/restud/rdaf022},
    doi = {10.1093/restud/rdaf022},
    journal = {The Review of Economic Studies},
    author = {Bordalo, Pedro and Conlon, John and Gennaioli, Nicola and Kwon, Spencer and Shleifer, Andrei},
    month = jan,
    year = {2026},
    pages = {250--285},
}

@article{bordalo_memory_2020,
    title = {Memory, attention, and choice},
    volume = {135},
    number = {3},
    url = {https://academic.oup.com/qje/article/135/3/1399/5824669},
    doi = {10.1093/qje/qjaa007},
    journal = {The Quarterly Journal of Economics},
    author = {Bordalo, Pedro and Gennaioli, Nicola and Shleifer, Andrei},
    month = aug,
    year = {2020},
    pages = {1399--1442},
}

@article{bordalo_salience_2012,
    title = {Salience theory of choice under risk},
    volume = {127},
    number = {3},
    url = {https://academic.oup.com/qje/article/127/3/1243/1922202},
    doi = {10.1093/qje/qjs018},
    journal = {The Quarterly Journal of Economics},
    author = {Bordalo, Pedro and Gennaioli, Nicola and Shleifer, Andrei},
    month = aug,
    year = {2012},
    pages = {1243--1285},
}

@article{bordalo_salience_2013,
    title = {Salience and consumer choice},
    volume = {121},
    number = {5},
    url = {https://www.jstor.org/stable/10.1086/673885},
    doi = {10.1086/673885},
    journal = {Journal of Political Economy},
    author = {Bordalo, Pedro and Gennaioli, Nicola and Shleifer, Andrei},
    month = oct,
    year = {2013},
    pages = {803--843},
}

@article{bordalo_stereotypes_2016,
    title = {Stereotypes},
    volume = {131},
    number = {4},
    url = {https://academic.oup.com/qje/article/131/4/1753/2468882},
    doi = {10.1093/qje/qjw029},
    journal = {The Quarterly Journal of Economics},
    author = {Bordalo, Pedro and Coffman, Katherine and Gennaioli, Nicola and Shleifer, Andrei},
    month = nov,
    year = {2016},
    pages = {1753--1794},
}

@article{caliskan_semantics_2017,
    title = {Semantics derived automatically from language corpora contain human-like biases},
    author = {Caliskan, Aylin and Bryson, Joanna J. and Narayanan, Arvind},
    journal = {Science},
    volume = {356},
    number = {6334},
    pages = {183--186},
    month = apr,
    year = {2017},
    doi = {10.1126/science.aal4230},
    url = {https://www.science.org/doi/10.1126/science.aal4230},
}

@article{chen_emergence_2023,
    title = {The emergence of economic rationality of {GPT}},
    author = {Chen, Yiting and Liu, Tracy Xiao and Shan, You and Zhong, Songfa},
    journal = {Proceedings of the National Academy of Sciences},
    volume = {120},
    number = {51},
    pages = {e2316205120},
    month = dec,
    year = {2023},
    doi = {10.1073/pnas.2316205120},
    url = {https://www.pnas.org/doi/10.1073/pnas.2316205120},
}

@article{cheung_large_2025,
    title = {Large language models show amplified cognitive biases in moral decision-making},
    author = {Cheung, Vanessa and Maier, Maximilian and Lieder, Falk},
    journal = {Proceedings of the National Academy of Sciences},
    volume = {122},
    number = {25},
    pages = {e2412015122},
    month = jun,
    year = {2025},
    doi = {10.1073/pnas.2412015122},
    url = {https://www.pnas.org/doi/10.1073/pnas.2412015122},
}

@techreport{cook_what_2026,
    title = {What do {LLMs} want?},
    author = {Cook, Thomas R. and Kazinnik, Sophia and Modig, Zach and Palmer, Nathan M.},
    type = {Finance and Economics Discussion Series},
    number = {2026-006},
    institution = {Board of Governors of the Federal Reserve System},
    month = jan,
    year = {2026},
    doi = {10.17016/FEDS.2026.006},
    url = {https://www.federalreserve.gov/econres/feds/what-do-llms-want.htm},
}

@article{dean_experimental_2023,
    title = {Experimental tests of rational inattention},
    author = {Dean, Mark and Neligh, Nathaniel},
    journal = {Journal of Political Economy},
    volume = {131},
    number = {12},
    pages = {3415--3461},
    month = dec,
    year = {2023},
    doi = {10.1086/725174},
    url = {https://www.journals.uchicago.edu/doi/full/10.1086/725174},
}

@misc{deepseek-ai_deepseek-v4_2026,
    title = {{DeepSeek}-{V4}: {Towards} highly efficient million-token context intelligence},
    author = {{DeepSeek-AI} and Xu, Anyi and Lin, Bangcai and Xue, Bing and Wang, Bingxuan and Xu, Bingzheng and Wu, Bochao and Zhang, Bowei and Lin, Chaofan and Dong, Chen and Ling, Chenchen and Lu, Chengda and Zhao, Chenggang and Deng, Chengqi and Hou, Chengyu and Xu, Chenhao and Shao, Chenze and Ruan, Chong and Sun, Conner and Dai, Damai and Guo, Daya and Yang, Dejian and Chen, Deli and Li, Donghao and Ji, Dongjie and Li, Erhang and Wei, Fang and Lin, Fangyun and Yuan, Fangzhou and Xia, Feiyu and Dai, Fucong and Hao, Guangbo and Chen, Guanting and Cao, Guoai and Meng, Guolai and Li, Guowei and Yu, Han and Zhang, Han and Xu, Hanwei and Li, Hao and Liang, Haofen and Zhang, Haoling and Luo, Haoming and Wei, Haoran and Yuan, Haotian and Zhang, Haowei and Luo, Haowen and Chen, Haoyu and Ji, Haozhe and Zhang, Hengqing and Ding, Honghui and Tang, Hongxuan and Cao, Huanqi and Gao, Huazuo and Qu, Hui and Zeng, Hui and Yang, J. and Zhu, J. Q. and Luo, Jia and Song, Jia and Yu, Jia and Huang, Jialiang and Cai, Jialu and Liang, Jian and Zhou, Jiangting and Ye, Jiasheng and Li, Jiashi and Xu, Jiaxin and Hu, Jiewen and Yang, Jieyu and Chen, Jin and Yan, Jin and Chen, Jingchang and Zhou, Jingli and Xiang, Jingting and Yuan, Jingyang and Cheng, Jingyuan and Zhou, Jingzi and Zhu, Jinhua and Yu, Jiping and Sun, Joseph and Ran, Jun and Jiang, Junguang and Qiu, Junjie and Li, Junlong and Zheng, Junmin and Song, Junxiao and Dong, Kai and Gao, Kaige and Guan, Kang and Zhou, Kexing and Huang, Kezhao and Yu, Kuai and Wang, Lean and Zhang, Lecong and Wang, Lei and Xia, Leyi and Zhang, Li and Zhao, Liang and Guo, Lihua and Luo, Lingxiao and Ma, Linwang and Zhu, Linyan and Wang, Litong and Cai, Liyu and Zhang, Liyue and Chen, Longhao and Di, M. S. and Xu, M. Y. and Mei, Max and Wang, Miaojun and Zhang, Mingchuan and Zhang, Minghua and Tang, Minghui and Li, Mingming and Zhou, Mingxu and Han, Minmin and Wang, Ning and Huang, Panpan and Wang, Panpan and Cong, Peixin and Wang, Peiyi and Zhang, Peng and Wang, Qiancheng and Zhu, Qihao and Li, Qingyang and Chen, Qinyu and Du, Qiushi and Jiang, Qiwei and Tian, Rui and Xu, Ruifan and Lu, Ruijie and Xu, Ruiling and Ge, Ruiqi and Zhang, Ruisong and Pan, Ruizhe and Wang, Runji and Chen, Runqian and Yin, Runqiu and Xu, Runxin and Shen, Ruomeng and Zhang, Ruoyu and Chen, Ruyi and Liu, S. H. and Lu, Shanghao and Sun, Shangmian and Zhou, Shangyan and Chen, Shanhuang and Cai, Shaofei and Nie, Shaoheng and Wu, Shaoqing and Chen, Shaoyuan and Hu, Shengding and Liu, Shengyu and Hu, Shiqiang and Ma, Shirong and Wang, Shiyu and Yu, Shuiping and Zhou, Shunfeng and Pan, Shuting and Yu, Shuying and Zhou, Songyang and Ni, Tao and Yun, Tao and Jin, Tian and Pei, Tian and Ye, Tian and Lin, Tianle and Ji, Tianran and Cui, Tianyi and Yue, Tianyuan and Yu, Tingting and Wang, Tun and Zhang, W. and Xiao, W. L. and Zeng, Wangding and An, Wei and Zhao, Weilin and Liu, Wen and Liang, Wenfeng and Pang, Wenjie and Luo, Wenjing and Yao, Wenjing and Gao, Wenjun and Yang, Wenkai and Huang, Wenlve and Hou, Wenqing and Zhang, Wentao and Ma, Wenting and Gao, Xi and He, Xiang and Wang, Xiangwen and Wang, Xianzu and Bi, Xiao and Liu, Xiaodong and Wang, Xiaohan and Chen, Xiaokang and Zhang, Xiaokang and Nie, Xiaotao and Sun, Xiaowen and Wang, Xiaoxiang and Cheng, Xin and Liu, Xin and Xie, Xin and Liu, Xingchao and Liu, Xingchen and Yu, Xingkai and Li, Xingyou and Yang, Xinyu and Zhang, Xinyu and Chen, Xu and Wang, Xuanyu and Su, Xuecheng and Chen, Xueyin and Lin, Xuheng and Fu, Xuwei and Yan, Y. C. and Wang, Y. Q. and Ma, Y. W. and Luo, Yanfeng and Zhang, Yang and Xu, Yanhong and Ma, Yanru and Huang, Yanwen and Li, Yao and Li, Yao and Xu, Yao and Zhao, Yao and Sun, Yaofeng and Wang, Yaohui and Qian, Yi and Shao, Yi and Yu, Yi and Zhang, Yichao and Ding, Yifan and Shi, Yifan and Wu, Yijia and Xiong, Yiliang and Ma, Yiling and He, Ying and Tang, Ying and Zhou, Ying and Luo, Yingjia and Zhong, Yinmin and Piao, Yishi and Wang, Yisong and Zhang, Yixiang and Chen, Yixiao and Tan, Yixuan and Wei, Yixuan and Ma, Yiyang and Liu, Yiyuan and Yang, Yonglun and Guo, Yongqiang and Wu, Yongtong and Wu, Yu and Li, YuKun and Cheng, Yuan and Ou, Yuan and Xu, Yuanfan and Li, Yuanhao and Wang, Yuduan and Yang, Yuehan and Xu, Yuer and Wu, Yuhan and Meng, Yuhao and Zou, Yuheng and Zha, Yukun and Xiong, Yunfan and Chen, Yupeng and Lin, Yuping and Cao, Yuqian and Wang, Yuqian and Zhang, Yushun and Yan, Yuting and Lin, Yutong and Gu, Yuxian and Luo, Yuxiang and You, Yuxiang and Liu, Yuxuan and Zhou, Yuxuan and Zhou, Yuyang and Huang, Yuzhen and Wu, Z. F. and Wang, Zehao and Zhao, Zehua and Ren, Zehui and Zhang, Zekai and Sha, Zhangli and Fu, Zhe and Ju, Zhe and Xu, Zhean and Xie, Zhenda and Zhang, Zhengyan and Gao, Zheren and Hao, Zhewen and Gou, Zhibin and Ma, Zhicheng and Yan, Zhigang and Shao, Zhihong and Huang, Zhixian and Chen, Zhixuan and Wu, Zhiyu and Ren, Zhizhou and Wu, Zhongyu and Li, Zhuoshu and Zhang, Zhuping and Xu, Zian and Wang, Zihao and Qu, Zihua and Gu, Zihui and Zhu, Zijia and Li, Zilin and Zhang, Zipeng and Xie, Ziwei and Gao, Ziyi and Wan, Ziyi and Pan, Zizheng and Yao, Zongqing},
    month = apr,
    year = {2026},
    publisher = {arXiv},
    note = {arXiv:2606.19348 [cs]},
    doi = {10.48550/arXiv.2606.19348},
    url = {https://arxiv.org/abs/2606.19348},
}

@article{dhar_consumer_2000,
    title = {Consumer choice between hedonic and utilitarian goods},
    author = {Dhar, Ravi and Wertenbroch, Klaus},
    journal = {Journal of Marketing Research},
    volume = {37},
    number = {1},
    pages = {60--71},
    month = feb,
    year = {2000},
    doi = {10.1509/jmkr.37.1.60.18718},
}

@inproceedings{echterhoff_cognitive_2024,
    title = {Cognitive bias in decision-making with {LLMs}},
    author = {Echterhoff, Jessica Maria and Liu, Yao and Alessa, Abeer and McAuley, Julian and He, Zexue},
    editor = {Al-Onaizan, Yaser and Bansal, Mohit and Chen, Yun-Nung},
    booktitle = {Findings of the {Association} for {Computational} {Linguistics}: {EMNLP} 2024},
    address = {Miami, Florida, USA},
    publisher = {Association for Computational Linguistics},
    pages = {12640--12653},
    month = nov,
    year = {2024},
    doi = {10.18653/v1/2024.findings-emnlp.739},
    url = {https://aclanthology.org/2024.findings-emnlp.739/},
}

@article{enke_cognitive_2023,
    title = {Cognitive uncertainty},
    author = {Enke, Benjamin and Graeber, Thomas},
    journal = {The Quarterly Journal of Economics},
    volume = {138},
    number = {4},
    pages = {2021--2067},
    year = {2023},
    doi = {10.1093/qje/qjad025},
    url = {https://academic.oup.com/qje/article/138/4/2021/7181327},
}

@incollection{gabaix_behavioral_2019,
    title = {Behavioral inattention},
    author = {Gabaix, Xavier},
    booktitle = {Handbook of {Behavioral} {Economics}: {Applications} and {Foundations} 1},
    editor = {Bernheim, B. Douglas and DellaVigna, Stefano and Laibson, David},
    volume = {2},
    pages = {261--343},
    publisher = {North-Holland},
    year = {2019},
    doi = {10.1016/bs.hesbe.2018.11.001},
    url = {https://www.sciencedirect.com/science/article/pii/S2352239918300216},
}

@article{gao_take_2025,
    title = {Take caution in using {LLMs} as human surrogates},
    author = {Gao, Yuan and Lee, Dokyun and Burtch, Gordon and Fazelpour, Sina},
    journal = {Proceedings of the National Academy of Sciences},
    volume = {122},
    number = {24},
    pages = {e2501660122},
    month = jun,
    year = {2025},
    doi = {10.1073/pnas.2501660122},
    url = {https://www.pnas.org/doi/10.1073/pnas.2501660122},
}

@inproceedings{geiger_finding_2024,
    title     = {Finding Alignments Between Interpretable Causal Variables and Distributed Neural Representations},
    author    = {Geiger, Atticus and Wu, Zhengxuan and Potts, Christopher and Icard, Thomas and Goodman, Noah},
    booktitle = {Proceedings of the Third Conference on Causal Learning and Reasoning},
    editor    = {Locatello, Francesco and Didelez, Vanessa},
    series    = {Proceedings of Machine Learning Research},
    volume    = {236},
    pages     = {160--187},
    month     = apr,
    year      = {2024},
    publisher = {PMLR},
    url       = {https://proceedings.mlr.press/v236/geiger24a.html}
}

@inproceedings{geva_transformer_2021,
    title     = {Transformer Feed-Forward Layers Are Key-Value Memories},
    author    = {Geva, Mor and Schuster, Roei and Berant, Jonathan and Levy, Omer},
    editor    = {Moens, Marie-Francine and Huang, Xuanjing and Specia, Lucia and Yih, Scott Wen-tau},
    booktitle = {Proceedings of the 2021 Conference on Empirical Methods in Natural Language Processing},
    month     = nov,
    year      = {2021},
    address   = {Online and Punta Cana, Dominican Republic},
    publisher = {Association for Computational Linguistics},
    pages     = {5484--5495},
    doi       = {10.18653/v1/2021.emnlp-main.446},
    url       = {https://aclanthology.org/2021.emnlp-main.446/}
}

@misc{glm-5-team_glm-5_2026,
    title = {{GLM}-5: {From} vibe coding to agentic engineering},
    author = {{GLM-5 Team} and Zeng, Aohan and Lv, Xin and Hou, Zhenyu and Du, Zhengxiao and Zheng, Qinkai and Chen, Bin and Yin, Da and Ge, Chendi and Huang, Chenghua and Xie, Chengxing and Zhu, Chenzheng and Yin, Congfeng and Wang, Cunxiang and Pan, Gengzheng and Zeng, Hao and Zhang, Haoke and Wang, Haoran and Chen, Huilong and Zhang, Jiajie and Jiao, Jian and Guo, Jiaqi and Wang, Jingsen and Du, Jingzhao and Wu, Jinzhu and Wang, Kedong and Li, Lei and Fan, Lin and Zhong, Lucen and Liu, Mingdao and Zhao, Mingming and Du, Pengfan and Dong, Qian and Lu, Rui and Li, Shuang and Cao, Shulin and Liu, Song and Jiang, Ting and Chen, Xiaodong and Zhang, Xiaohan and Huang, Xuancheng and Dong, Xuezhen and Xu, Yabo and Wei, Yao and An, Yifan and Niu, Yilin and Zhu, Yitong and Wen, Yuanhao and Cen, Yukuo and Bai, Yushi and Qiao, Zhongpei and Wang, Zihan and Wang, Zikang and Zhu, Zilin and Liu, Ziqiang and Li, Zixuan and Wang, Bojie and Wen, Bosi and Huang, Can and Cai, Changpeng and Yu, Chao and Li, Chen and Hu, Chengwei and Zhang, Chenhui and Zhang, Dan and Lin, Daoyan and Yang, Dayong and Wang, Di and Ai, Ding and Zhu, Erle and Yi, Fangzhou and Chen, Feiyu and Wen, Guohong and Sun, Hailong and Zhao, Haisha and Hu, Haiyi and Zhang, Hanchen and Liu, Hanrui and Zhang, Hanyu and Peng, Hao and Tai, Hao and Zhang, Haobo and Liu, He and Wang, Hongwei and Yan, Hongxi and Ge, Hongyu and Liu, Huan and Chu, Huanpeng and Zhao, Jia'ni and Wang, Jiachen and Zhao, Jiajing and Ren, Jiamin and Wang, Jiapeng and Zhang, Jiaxin and Gui, Jiayi and Zhao, Jiayue and Li, Jijie and An, Jing and Li, Jing and Yuan, Jingwei and Du, Jinhua and Liu, Jinxin and Zhi, Junkai and Duan, Junwen and Zhou, Kaiyue and Wei, Kangjian and Wang, Ke and Luo, Keyun and Zhang, Laiqiang and Sha, Leigang and Xu, Liang and Wu, Lindong and Ding, Lintao and Chen, Lu and Li, Minghao and Lin, Nianyi and Ta, Pan and Zou, Qiang and Song, Rongjun and Yang, Ruiqi and Tu, Shangqing and Yang, Shangtong and Wu, Shaoxiang and Zhang, Shengyan and Li, Shijie and Li, Shuang and Fan, Shuyi and Qin, Wei and Tian, Wei and Zhang, Weining and Yu, Wenbo and Liang, Wenjie and Kuang, Xiang and Cheng, Xiangmeng and Li, Xiangyang and Yan, Xiaoquan and Hu, Xiaowei and Ling, Xiaoying and Fan, Xing and Xia, Xingye and Zhang, Xinyuan and Zhang, Xinze and Pan, Xirui and Zou, Xu and Zhang, Xunkai and Liu, Yadi and Wu, Yandong and Li, Yanfu and Wang, Yidong and Zhu, Yifan and Tan, Yijun and Zhou, Yilin and Pan, Yiming and Zhang, Ying and Su, Yinpei and Geng, Yipeng and Yan, Yong and Tan, Yonglin and Bi, Yuean and Shen, Yuhan and Yang, Yuhao and Li, Yujiang and Liu, Yunan and Wang, Yunqing and Li, Yuntao and Wu, Yurong and Zhang, Yutao and Duan, Yuxi and Zhang, Yuxuan and Liu, Zezhen and Jiang, Zhengtao and Yan, Zhenhe and Zhang, Zheyu and Wei, Zhixiang and Chen, Zhuo and Feng, Zhuoer and Yao, Zijun and Chai, Ziwei and Wang, Ziyuan and Zhang, Zuzhou and Xu, Bin and Huang, Minlie and Wang, Hongning and Li, Juanzi and Dong, Yuxiao and Tang, Jie},
    month = feb,
    year = {2026},
    publisher = {arXiv},
    note = {arXiv:2602.15763 [cs]},
    doi = {10.48550/arXiv.2602.15763},
    url = {https://arxiv.org/abs/2602.15763},
}

@misc{google_deepmind_gemini_2025,
    title = {{Gemini 3 Flash} model card},
    author = {{Google DeepMind}},
    howpublished = {Model card},
    month = dec,
    year = {2025},
    url = {https://storage.googleapis.com/deepmind-media/Model-Cards/Gemini-3-Flash-Model-Card.pdf},
    urldate = {2026-08-29},
}

@misc{gui_challenge_2023,
    title = {The challenge of using {LLMs} to simulate human behavior: {A} causal inference perspective},
    author = {Gui, George and Toubia, Olivier},
    publisher = {arXiv},
    month = nov,
    year = {2025},
    doi = {10.48550/arXiv.2312.15524},
    url = {https://arxiv.org/abs/2312.15524},
    note = {arXiv:2312.15524 [cs].},
}

@inproceedings{gupta_bias_2024,
    title = {Bias runs deep: Implicit reasoning biases in persona-assigned {LLMs}},
    author = {Gupta, Shashank and Shrivastava, Vaishnavi and Deshpande, Ameet and Kalyan, Ashwin and Clark, Peter and Sabharwal, Ashish and Khot, Tushar},
    booktitle = {The Twelfth International Conference on Learning Representations},
    year = {2024},
    url = {https://openreview.net/forum?id=kGteeZ18Ir},
}

@article{hagendorff_human-like_2023,
    title = {Human-like intuitive behavior and reasoning biases emerged in large language models but disappeared in {ChatGPT}},
    author = {Hagendorff, Thilo and Fabi, Sarah and Kosinski, Michal},
    journal = {Nature Computational Science},
    volume = {3},
    number = {10},
    pages = {833--838},
    month = oct,
    year = {2023},
    doi = {10.1038/s43588-023-00527-x},
    url = {https://www.nature.com/articles/s43588-023-00527-x},
}

@article{hamilton_low_2013,
    title = {Low prices are just the beginning: {Price} image in retail management},
    author = {Hamilton, Ryan and Chernev, Alexander},
    journal = {Journal of Marketing},
    volume = {77},
    number = {6},
    pages = {1--20},
    month = nov,
    year = {2013},
    doi = {10.1509/jm.08.0204},
}

@article{helion_gift_2014,
    title = {Gift cards and mental accounting: {Green}-lighting hedonic spending},
    author = {Helion, Chelsea and Gilovich, Thomas},
    journal = {Journal of Behavioral Decision Making},
    volume = {27},
    number = {4},
    pages = {386--393},
    month = oct,
    year = {2014},
    doi = {10.1002/bdm.1813},
    url = {https://onlinelibrary.wiley.com/doi/10.1002/bdm.1813},
}

@article{hirschman_hedonic_1982,
    title = {Hedonic consumption: {Emerging} concepts, methods and propositions},
    author = {Hirschman, Elizabeth C. and Holbrook, Morris B.},
    journal = {Journal of Marketing},
    volume = {46},
    number = {3},
    pages = {92--101},
    month = jul,
    year = {1982},
    doi = {10.1177/002224298204600314},
    url = {https://www.jstor.org/stable/1251707},
}

@misc{hullman_this_2026,
    title = {This human study did not involve human subjects: {Validating} {LLM} simulations as behavioral evidence},
    author = {Hullman, Jessica and Broska, David and Sun, Huaman and Shaw, Aaron},
    publisher = {arXiv},
    month = feb,
    year = {2026},
    doi = {10.48550/arXiv.2602.15785},
    url = {https://arxiv.org/abs/2602.15785},
    note = {arXiv:2602.15785 [cs]},
}

@unpublished{imas_agentic_2025,
    title = {Agentic interactions},
    author = {Imas, Alex and Lee, Kevin and Misra, Sanjog},
    note = {{SSRN} Working Paper 5875162},
    month = dec,
    year = {2025},
    doi = {10.2139/ssrn.5875162},
    url = {https://papers.ssrn.com/abstract=5875162},
}

@article{itzhak_instructed_2024,
    title = {Instructed to bias: {Instruction}-tuned language models exhibit emergent cognitive bias},
    author = {Itzhak, Itay and Stanovsky, Gabriel and Rosenfeld, Nir and Belinkov, Yonatan},
    journal = {Transactions of the Association for Computational Linguistics},
    volume = {12},
    pages = {771--785},
    year = {2024},
    publisher = {MIT Press},
    address = {Cambridge, MA},
    doi = {10.1162/tacl_a_00673},
    url = {https://aclanthology.org/2024.tacl-1.43/},
}

@article{jiang_investor_2025,
    title = {Investor memory and biased beliefs: {Evidence} from the field},
    author = {Jiang, Zhengyang and Liu, Hongqi and Peng, Cameron and Yan, Hongjun},
    journal = {The Quarterly Journal of Economics},
    volume = {140},
    number = {4},
    pages = {2749--2804},
    year = {2025},
    doi = {10.1093/qje/qjaf035},
    url = {https://doi.org/10.1093/qje/qjaf035},
}

@inproceedings{jones_capturing_2022,
    title     = {Capturing Failures of Large Language Models via Human Cognitive Biases},
    author    = {Jones, Erik and Steinhardt, Jacob},
    booktitle = {Advances in Neural Information Processing Systems},
    editor    = {Koyejo, S. and Mohamed, S. and Agarwal, A. and Belgrave, D. and Cho, K. and Oh, A.},
    volume    = {35},
    pages     = {11785--11799},
    year      = {2022},
    publisher = {Curran Associates, Inc.},
    url       = {https://proceedings.neurips.cc/paper_files/paper/2022/hash/4d13b2d99519c5415661dad44ab7edcd-Abstract-Conference.html}
}

@book{kahana_foundations_2012,
    title = {Foundations of Human Memory},
    author = {Kahana, Michael J.},
    publisher = {Oxford University Press},
    address = {New York},
    year = {2012},
    isbn = {978-0-19-533324-4},
}

@misc{karvonen_robustly_2025,
    title = {Robustly improving {LLM} fairness in realistic settings via interpretability},
    author = {Karvonen, Adam and Marks, Samuel},
    publisher = {arXiv},
    month = jun,
    year = {2025},
    note = {arXiv:2506.10922 [cs]},
    doi = {10.48550/arXiv.2506.10922},
    url = {https://arxiv.org/abs/2506.10922},
}

@article{khan_licensing_2006,
    title = {Licensing effect in consumer choice},
    author = {Khan, Uzma and Dhar, Ravi},
    journal = {Journal of Marketing Research},
    volume = {43},
    number = {2},
    pages = {259--266},
    month = may,
    year = {2006},
    doi = {10.1509/jmkr.43.2.259},
}

@article{koschmann_retailer_2018,
    title = {Retailer categorization: {How} store-format price image influences expected prices and consumer choices},
    author = {Koschmann, Anthony and Isaac, Mathew S.},
    journal = {Journal of Retailing},
    volume = {94},
    number = {4},
    pages = {364--379},
    month = dec,
    year = {2018},
    doi = {10.1016/j.jretai.2018.08.001},
    url = {https://www.sciencedirect.com/science/article/pii/S002243591830037X},
}

@article{levav_emotional_2009,
    title = {Emotional accounting: {How} feelings about money influence consumer choice},
    author = {Levav, Jonathan and McGraw, A. Peter},
    journal = {Journal of Marketing Research},
    volume = {46},
    number = {1},
    pages = {66--80},
    month = feb,
    year = {2009},
    doi = {10.1509/jmkr.46.1.66},
}

@article{li_frontiers_2024,
    title = {Frontiers: {Determining} the validity of large language models for automated perceptual analysis},
    author = {Li, Peiyao and Castelo, Noah and Katona, Zsolt and Sarvary, Miklos},
    journal = {Marketing Science},
    volume = {43},
    number = {2},
    pages = {254--266},
    month = jan,
    year = {2024},
    doi = {10.1287/mksc.2023.0454},
    url = {https://pubsonline.informs.org/doi/abs/10.1287/mksc.2023.0454},
}

@article{liberman_role_1998,
    title = {The role of feasibility and desirability considerations in near and distant future decisions: {A} test of temporal construal theory},
    author = {Liberman, Nira and Trope, Yaacov},
    journal = {Journal of Personality and Social Psychology},
    volume = {75},
    number = {1},
    pages = {5--18},
    month = jul,
    year = {1998},
    doi = {10.1037/0022-3514.75.1.5},
}

@article{liew_appropriacy_2016,
    title = {The appropriacy of averaging in the study of context effects},
    author = {Liew, Shi Xian and Howe, Piers D. L. and Little, Daniel R.},
    journal = {Psychonomic Bulletin \& Review},
    volume = {23},
    number = {5},
    pages = {1639--1646},
    month = oct,
    year = {2016},
    doi = {10.3758/s13423-016-1032-7},
    url = {https://link.springer.com/10.3758/s13423-016-1032-7},
}

@misc{lior_comparing_2026,
    title = {Comparing the framing effect in humans and {LLMs} on naturally occurring texts},
    author = {Lior, Gili and Naccache, Liron and Stanovsky, Gabriel},
    publisher = {arXiv},
    month = jan,
    year = {2026},
    note = {arXiv:2502.17091 [cs]},
    doi = {10.48550/arXiv.2502.17091},
    url = {https://arxiv.org/abs/2502.17091},
}

@article{mackowiak_rational_2023,
    title = {Rational inattention: {A} review},
    author = {Maćkowiak, Bartosz and Matějka, Filip and Wiederholt, Mirko},
    journal = {Journal of Economic Literature},
    volume = {61},
    number = {1},
    pages = {226--273},
    month = mar,
    year = {2023},
    doi = {10.1257/jel.20211524},
    url = {https://www.aeaweb.org/articles?id=10.1257/jel.20211524},
}

@inproceedings{malberg_comprehensive_2025,
    title = {A comprehensive evaluation of cognitive biases in {LLMs}},
    author = {Malberg, Simon and Poletukhin, Roman and Schuster, Carolin M. and Groh, Georg},
    editor = {H{\"a}m{\"a}l{\"a}inen, Mika and {\"O}hman, Emily and Bizzoni, Yuri and Miyagawa, So and Alnajjar, Khalid},
    booktitle = {Proceedings of the 5th {International} {Conference} on {Natural} {Language} {Processing} for {Digital} {Humanities}},
    address = {Albuquerque, USA},
    publisher = {Association for Computational Linguistics},
    pages = {578--613},
    month = may,
    year = {2025},
    doi = {10.18653/v1/2025.nlp4dh-1.50},
    url = {https://aclanthology.org/2025.nlp4dh-1.50/},
}

@techreport{manning_general_2025,
    title = {General social agents},
    author = {Manning, Benjamin S. and Horton, John J.},
    type = {Working Paper},
    number = {34937},
    institution = {National Bureau of Economic Research},
    month = mar,
    year = {2026},
    doi = {10.3386/w34937},
    url = {https://www.nber.org/papers/w34937},
}

@book{manski_identification_1995,
    title = {Identification Problems in the Social Sciences},
    author = {Manski, Charles F.},
    publisher = {Harvard University Press},
    address = {Cambridge, MA},
    year = {1995},
    isbn = {978-0-674-44283-2},
}

@misc{marks_persona_2026,
    title = {The persona selection model: {Why} {AI} assistants might behave like humans},
    author = {Marks, Sam and Lindsey, Jack and Olah, Christopher},
    year = {2026},
    month = feb,
    howpublished = {Anthropic Alignment Science Blog},
    url = {https://alignment.anthropic.com/2026/psm/},
    urldate = {2026-08-29},
}

@article{mccoy_embers_2023,
    title = {Embers of autoregression show how large language models are shaped by the problem they are trained to solve},
    author = {McCoy, R. Thomas and Yao, Shunyu and Friedman, Dan and Hardy, Mathew D. and Griffiths, Thomas L.},
    journal = {Proceedings of the National Academy of Sciences},
    volume = {121},
    number = {41},
    pages = {e2322420121},
    month = oct,
    year = {2024},
    doi = {10.1073/pnas.2322420121},
    url = {https://www.pnas.org/doi/10.1073/pnas.2322420121},
}

@article{okada_justification_2005,
    title = {Justification effects on consumer choice of hedonic and utilitarian goods},
    author = {Okada, Erica Mina},
    journal = {Journal of Marketing Research},
    volume = {42},
    number = {1},
    pages = {43--53},
    month = feb,
    year = {2005},
    doi = {10.1509/jmkr.42.1.43.56889},
    url = {https://doi.org/10.1509/jmkr.42.1.43.56889},
}

@misc{openai_gpt_2026,
    title = {{GPT-5.4 Thinking} system card},
    author = {{OpenAI}},
    howpublished = {System card},
    month = mar,
    year = {2026},
    url = {https://deploymentsafety.openai.com/gpt-5-4-thinking/gpt-5-4-thinking.pdf},
    urldate = {2026-08-29},
}

@misc{park_generative_2024,
    title = {{LLM} agents grounded in self-reports enable general-purpose simulation of individuals},
    author = {Park, Joon Sung and Zou, Carolyn Q. and Kamphorst, Jonne and Egan, Niles and Shaw, Aaron and Hill, Benjamin Mako and Cai, Carrie and Morris, Meredith Ringel and Liang, Percy and Willer, Robb and Bernstein, Michael S.},
    publisher = {arXiv},
    month = jun,
    year = {2026},
    note = {arXiv:2411.10109 [cs]},
    doi = {10.48550/arXiv.2411.10109},
    url = {https://arxiv.org/abs/2411.10109},
}

@inproceedings{pawar_presumed_2025,
    title = {Presumed cultural identity: How names shape {LLM} responses},
    author = {Pawar, Siddhesh Milind and Arora, Arnav and Kaffee, Lucie-Aim{\'e}e and Augenstein, Isabelle},
    booktitle = {Findings of the Association for Computational Linguistics: EMNLP 2025},
    editor = {Christodoulopoulos, Christos and Chakraborty, Tanmoy and Rose, Carolyn and Peng, Violet},
    publisher = {Association for Computational Linguistics},
    address = {Suzhou, China},
    pages = {22147--22172},
    month = nov,
    year = {2025},
    doi = {10.18653/v1/2025.findings-emnlp.1207},
    url = {https://aclanthology.org/2025.findings-emnlp.1207/},
}

@article{plott_willingness_2005,
    title = {The willingness to pay--willingness to accept gap, the ``endowment effect,'' subject misconceptions, and experimental procedures for eliciting valuations},
    author = {Plott, Charles R. and Zeiler, Kathryn},
    journal = {American Economic Review},
    volume = {95},
    number = {3},
    pages = {530--545},
    month = jun,
    year = {2005},
    doi = {10.1257/0002828054201387},
    url = {https://www.aeaweb.org/articles?id=10.1257/0002828054201387},
}

@article{reutskaja_search_2011,
    title = {Search dynamics in consumer choice under time pressure: {An} eye-tracking study},
    author = {Reutskaja, Elena and Nagel, Rosemarie and Camerer, Colin F. and Rangel, Antonio},
    journal = {American Economic Review},
    volume = {101},
    number = {2},
    pages = {900--926},
    month = apr,
    year = {2011},
    doi = {10.1257/aer.101.2.900},
    url = {https://www.aeaweb.org/articles?id=10.1257/aer.101.2.900},
}

@article{rosen_hedonic_1974,
    title = {Hedonic prices and implicit markets: {Product} differentiation in pure competition},
    author = {Rosen, Sherwin},
    journal = {Journal of Political Economy},
    volume = {82},
    number = {1},
    pages = {34--55},
    year = {1974},
    doi = {10.1086/260169},
    url = {https://www.jstor.org/stable/1830899},
}

@inproceedings{ross_llm_2024,
    title = {{LLM} economicus? {Mapping} the behavioral biases of {LLMs} via utility theory},
    author = {Ross, Jillian and Kim, Yoon and Lo, Andrew W.},
    booktitle = {First Conference on Language Modeling},
    year = {2024},
    url = {https://openreview.net/forum?id=Rx3wC8sCTJ},
}

@techreport{schumacher_agentic_2025,
    title       = {The agentic commerce opportunity: {How} {AI} agents are ushering in a new era for consumers and merchants},
    author      = {Schumacher, Katharina and Roberts, Roger and Giebel, Katharina},
    institution = {McKinsey \& Company},
    type        = {Report},
    month       = oct,
    year        = {2025},
    url         = {https://www.mckinsey.com/capabilities/quantumblack/our-insights/the-agentic-commerce-opportunity-how-ai-agents-are-ushering-in-a-new-era-for-consumers-and-merchants},
    urldate     = {2026-05-24},
}

@inproceedings{sclar_quantifying_2024,
    title = {Quantifying language models' sensitivity to spurious features in prompt design or: How {I} learned to start worrying about prompt formatting},
    author = {Sclar, Melanie and Choi, Yejin and Tsvetkov, Yulia and Suhr, Alane},
    booktitle = {The Twelfth International Conference on Learning Representations},
    year = {2024},
    url = {https://openreview.net/forum?id=RIu5lyNXjT},
}

@misc{shubatt_tradeoffs_2026,
    title = {Tradeoffs and comparison complexity},
    author = {Shubatt, Cassidy and Yang, Jeffrey},
    publisher = {arXiv},
    month = mar,
    year = {2026},
    doi = {10.48550/arXiv.2401.17578},
    url = {https://arxiv.org/abs/2401.17578},
    note = {arXiv:2401.17578 [econ]},
}

@article{stanovich_individual_2000,
    title = {Individual differences in reasoning: {Implications} for the rationality debate?},
    author = {Stanovich, Keith E. and West, Richard F.},
    journal = {Behavioral and Brain Sciences},
    volume = {23},
    number = {5},
    pages = {645--665},
    month = oct,
    year = {2000},
    doi = {10.1017/S0140525X00003435},
}

@article{stella_using_2023,
    title = {Using cognitive psychology to understand {GPT}-like models needs to extend beyond human biases},
    author = {Stella, Massimo and Hills, Thomas T. and Kenett, Yoed N.},
    journal = {Proceedings of the National Academy of Sciences},
    volume = {120},
    number = {43},
    pages = {e2312911120},
    month = oct,
    year = {2023},
    doi = {10.1073/pnas.2312911120},
    url = {https://www.pnas.org/doi/10.1073/pnas.2312911120},
}

@misc{stripe_agentic_nodate,
    title = {{Agentic Commerce Protocol}},
    author = {{Stripe}},
    howpublished = {Web page},
    month = sep,
    year = {2025},
    note = {Open standard codeveloped by Stripe and OpenAI, announced 29 September 2025},
    url = {https://www.agenticcommerce.dev/},
    urldate = {2026-08-29},
}

@misc{sun_persona_2026,
    title = {Persona vectors in games: {Measuring} and steering strategies via activation vectors},
    author = {Sun, Johnathan and Zhang, Andrew},
    year = {2026},
    month = mar,
    publisher = {arXiv},
    note = {arXiv:2603.21398 [cs]},
    doi = {10.48550/arXiv.2603.21398},
    url = {https://arxiv.org/abs/2603.21398},
}

@misc{team_gemma_2025,
    title = {Gemma 3 technical report},
    author = {{Gemma Team} and Kamath, Aishwarya and Ferret, Johan and Pathak, Shreya and Vieillard, Nino and Merhej, Ramona and Perrin, Sarah and Matejovicova, Tatiana and Ramé, Alexandre and Rivière, Morgane and Rouillard, Louis and Mesnard, Thomas and Cideron, Geoffrey and Grill, Jean-bastien and Ramos, Sabela and Yvinec, Edouard and Casbon, Michelle and Pot, Etienne and Penchev, Ivo and Liu, Gaël and Visin, Francesco and Kenealy, Kathleen and Beyer, Lucas and Zhai, Xiaohai and Tsitsulin, Anton and Busa-Fekete, Robert and Feng, Alex and Sachdeva, Noveen and Coleman, Benjamin and Gao, Yi and Mustafa, Basil and Barr, Iain and Parisotto, Emilio and Tian, David and Eyal, Matan and Cherry, Colin and Peter, Jan-Thorsten and Sinopalnikov, Danila and Bhupatiraju, Surya and Agarwal, Rishabh and Kazemi, Mehran and Malkin, Dan and Kumar, Ravin and Vilar, David and Brusilovsky, Idan and Luo, Jiaming and Steiner, Andreas and Friesen, Abe and Sharma, Abhanshu and Sharma, Abheesht and Gilady, Adi Mayrav and Goedeckemeyer, Adrian and Saade, Alaa and Feng, Alex and Kolesnikov, Alexander and Bendebury, Alexei and Abdagic, Alvin and Vadi, Amit and György, András and Pinto, André Susano and Das, Anil and Bapna, Ankur and Miech, Antoine and Yang, Antoine and Paterson, Antonia and Shenoy, Ashish and Chakrabarti, Ayan and Piot, Bilal and Wu, Bo and Shahriari, Bobak and Petrini, Bryce and Chen, Charlie and Lan, Charline Le and Choquette-Choo, Christopher A. and Carey, C. J. and Brick, Cormac and Deutsch, Daniel and Eisenbud, Danielle and Cattle, Dee and Cheng, Derek and Paparas, Dimitris and Sreepathihalli, Divyashree Shivakumar and Reid, Doug and Tran, Dustin and Zelle, Dustin and Noland, Eric and Huizenga, Erwin and Kharitonov, Eugene and Liu, Frederick and Amirkhanyan, Gagik and Cameron, Glenn and Hashemi, Hadi and Klimczak-Plucińska, Hanna and Singh, Harman and Mehta, Harsh and Lehri, Harshal Tushar and Hazimeh, Hussein and Ballantyne, Ian and Szpektor, Idan and Nardini, Ivan and Pouget-Abadie, Jean and Chan, Jetha and Stanton, Joe and Wieting, John and Lai, Jonathan and Orbay, Jordi and Fernandez, Joseph and Newlan, Josh and Ji, Ju-yeong and Singh, Jyotinder and Black, Kat and Yu, Kathy and Hui, Kevin and Vodrahalli, Kiran and Greff, Klaus and Qiu, Linhai and Valentine, Marcella and Coelho, Marina and Ritter, Marvin and Hoffman, Matt and Watson, Matthew and Chaturvedi, Mayank and Moynihan, Michael and Ma, Min and Babar, Nabila and Noy, Natasha and Byrd, Nathan and Roy, Nick and Momchev, Nikola and Chauhan, Nilay and Sachdeva, Noveen and Bunyan, Oskar and Botarda, Pankil and Caron, Paul and Rubenstein, Paul Kishan and Culliton, Phil and Schmid, Philipp and Sessa, Pier Giuseppe and Xu, Pingmei and Stanczyk, Piotr and Tafti, Pouya and Shivanna, Rakesh and Wu, Renjie and Pan, Renke and Rokni, Reza and Willoughby, Rob and Vallu, Rohith and Mullins, Ryan and Jerome, Sammy and Smoot, Sara and Girgin, Sertan and Iqbal, Shariq and Reddy, Shashir and Sheth, Shruti and Põder, Siim and Bhatnagar, Sijal and Panyam, Sindhu Raghuram and Eiger, Sivan and Zhang, Susan and Liu, Tianqi and Yacovone, Trevor and Liechty, Tyler and Kalra, Uday and Evci, Utku and Misra, Vedant and Roseberry, Vincent and Feinberg, Vlad and Kolesnikov, Vlad and Han, Woohyun and Kwon, Woosuk and Chen, Xi and Chow, Yinlam and Zhu, Yuvein and Wei, Zichuan and Egyed, Zoltan and Cotruta, Victor and Giang, Minh and Kirk, Phoebe and Rao, Anand and Black, Kat and Babar, Nabila and Lo, Jessica and Moreira, Erica and Martins, Luiz Gustavo and Sanseviero, Omar and Gonzalez, Lucas and Gleicher, Zach and Warkentin, Tris and Mirrokni, Vahab and Senter, Evan and Collins, Eli and Barral, Joelle and Ghahramani, Zoubin and Hadsell, Raia and Matias, Yossi and Sculley, D. and Petrov, Slav and Fiedel, Noah and Shazeer, Noam and Vinyals, Oriol and Dean, Jeff and Hassabis, Demis and Kavukcuoglu, Koray and Farabet, Clement and Buchatskaya, Elena and Alayrac, Jean-Baptiste and Anil, Rohan and Lepikhin, Dmitry and Borgeaud, Sebastian and Bachem, Olivier and Joulin, Armand and Andreev, Alek and Hardin, Cassidy and Dadashi, Robert and Hussenot, Léonard},
    month = mar,
    year = {2025},
    publisher = {arXiv},
    note = {arXiv:2503.19786 [cs]},
    doi = {10.48550/arXiv.2503.19786},
    url = {https://arxiv.org/abs/2503.19786},
}

@article{thaler_mental_1999,
    title = {Mental accounting matters},
    author = {Thaler, Richard H.},
    journal = {Journal of Behavioral Decision Making},
    volume = {12},
    number = {3},
    pages = {183--206},
    month = sep,
    year = {1999},
    doi = {10.1002/(SICI)1099-0771(199909)12:3<183::AID-BDM318>3.0.CO;2-F},
}

@inproceedings{vaswani_attention_2017,
    title     = {Attention Is All You Need},
    author    = {Vaswani, Ashish and Shazeer, Noam and Parmar, Niki and Uszkoreit, Jakob and Jones, Llion and Gomez, Aidan N. and Kaiser, {\L}ukasz and Polosukhin, Illia},
    booktitle = {Advances in Neural Information Processing Systems},
    editor    = {Guyon, I. and Von Luxburg, U. and Bengio, S. and Wallach, H. and Fergus, R. and Vishwanathan, S. and Garnett, R.},
    volume    = {30},
    pages     = {5998--6008},
    year      = {2017},
    publisher = {Curran Associates, Inc.},
    url       = {https://proceedings.neurips.cc/paper/2017/hash/3f5ee243547dee91fbd053c1c4a845aa-Abstract.html}
}

@inproceedings{vig_causal_2020,
    title = {Investigating gender bias in language models using causal mediation analysis},
    author = {Vig, Jesse and Gehrmann, Sebastian and Belinkov, Yonatan and Qian, Sharon and Nevo, Daniel and Singer, Yaron and Shieber, Stuart},
    booktitle = {Advances in Neural Information Processing Systems 33},
    editor = {Larochelle, H. and Ranzato, M. and Hadsell, R. and Balcan, M. F. and Lin, H.},
    year = {2020},
    pages = {12388--12401},
    url = {https://proceedings.neurips.cc/paper/2020/hash/92650b2e92217715fe312e6fa7b90d82-Abstract.html},
}

@article{wakefield_situational_2003,
    title = {Situational price sensitivity: the role of consumption occasion, social context and income},
    author = {Wakefield, Kirk L. and Inman, J. Jeffrey},
    journal = {Journal of Retailing},
    volume = {79},
    number = {4},
    pages = {199--212},
    year = {2003},
    doi = {10.1016/j.jretai.2003.09.004},
    url = {https://www.sciencedirect.com/science/article/pii/S0022435903000551},
}

@misc{wang_rethinking_2026,
    title = {Rethinking prospect theory for {LLMs}: {Revealing} the instability of decision-making under epistemic uncertainty},
    author = {Wang, Rui and Lin, Qihan and Liu, Jiayu and Zong, Qing and Zheng, Tianshi and Guo, Dadi and Shi, Haochen and Han, Peixuan and Wang, Weiqi and Song, Yangqiu},
    publisher = {arXiv},
    month = jul,
    year = {2026},
    note = {arXiv:2508.08992 [cs]},
    doi = {10.48550/arXiv.2508.08992},
    url = {https://arxiv.org/abs/2508.08992},
}

@misc{buchanan_innate_2026,
    title = {The Innate Economic Preferences of Language Models},
    url = {http://arxiv.org/abs/2607.26288},
    doi = {10.48550/arXiv.2607.26288},
    urldate = {2026-08-30},
    publisher = {arXiv},
    author = {Buchanan, Joy and Foster, Joshua},
    month = jul,
    year = {2026},
    note = {arXiv:2607.26288 [econ]},
}

@inproceedings{xie_explanation_2022,
    title = {An explanation of in-context learning as implicit {Bayesian} inference},
    author = {Xie, Sang Michael and Raghunathan, Aditi and Liang, Percy and Ma, Tengyu},
    booktitle = {The Tenth International Conference on Learning Representations},
    year = {2022},
    url = {https://openreview.net/forum?id=RdJVFCHjUMI},
}

@misc{yang_qwen3_2025,
    title = {Qwen3 technical report},
    author = {Yang, An and Li, Anfeng and Yang, Baosong and Zhang, Beichen and Hui, Binyuan and Zheng, Bo and Yu, Bowen and Gao, Chang and Huang, Chengen and Lv, Chenxu and Zheng, Chujie and Liu, Dayiheng and Zhou, Fan and Huang, Fei and Hu, Feng and Ge, Hao and Wei, Haoran and Lin, Huan and Tang, Jialong and Yang, Jian and Tu, Jianhong and Zhang, Jianwei and Yang, Jianxin and Yang, Jiaxi and Zhou, Jing and Zhou, Jingren and Lin, Junyang and Dang, Kai and Bao, Keqin and Yang, Kexin and Yu, Le and Deng, Lianghao and Li, Mei and Xue, Mingfeng and Li, Mingze and Zhang, Pei and Wang, Peng and Zhu, Qin and Men, Rui and Gao, Ruize and Liu, Shixuan and Luo, Shuang and Li, Tianhao and Tang, Tianyi and Yin, Wenbiao and Ren, Xingzhang and Wang, Xinyu and Zhang, Xinyu and Ren, Xuancheng and Fan, Yang and Su, Yang and Zhang, Yichang and Zhang, Yinger and Wan, Yu and Liu, Yuqiong and Wang, Zekun and Cui, Zeyu and Zhang, Zhenru and Zhou, Zhipeng and Qiu, Zihan},
    month = may,
    year = {2025},
    publisher = {arXiv},
    note = {arXiv:2505.09388 [cs]},
    doi = {10.48550/arXiv.2505.09388},
    url = {https://arxiv.org/abs/2505.09388},
}

\appendix
\section{Experimental Details}
\label{app:design}

\subsection{Product pairs}
\label{app:pairs}

Each product's five options form a labeled quality ladder (budget, economy,
midrange, premium, luxury), with attributes drawn from a fixed per-product
schema (e.g., blade material, handle, and edge for chef's knives) so that
adjacent tiers differ in attribute values rather than description
style. Binary pairs use adjacent tiers (1--2 through 4--5). Prices are taken as
generated, and realized premium/standard price ratios span 1.4--3.5$\times$
(median 2.2$\times$). \Cref{tab:pair_examples} shows two example pairs.

% --- end inlined: tab_pair_examples.tex ---

\subsection{Cue texts}
\label{app:cues}

Each of the five cue dimensions (\Cref{tab:cues}) has one
paired hedonic/functional cue per product, so a framed trial differs from
its counterpart only in the cue sentence. The payment source, justification,
and reasoning frame cues use a fixed template with a product-specific stem, while the purchase trigger and retail architecture cues are instantiated per
product by Claude Sonnet~4.6 so that they read naturally for the specific
product (e.g., staff notes on ``blade feel'' for the chef's knife). Cues
never mention price, quality, or either option's attributes.
\Cref{tab:cue_templates} shows all five dimensions instantiated for one
product.

\subsection{Prompt template and trial structure}
\label{app:prompt}
\label{app:trials}

Cues are written from the shopper's perspective. The full template is

\begin{quote}\small\ttfamily
\{cue\_text\} Here are two options.\\[4pt]
**Option A: \{a\_name\}**\\
\{a\_description\}\\
Price: \$\{a\_price\}\\[4pt]
**Option B: \{b\_name\}**\\
\{b\_description\}\\
Price: \$\{b\_price\}\\[4pt]
Which option should I choose? Answer with only ``A'' or ``B''.
\end{quote}

Each of the 800 (product, tier pair, order) combinations appears under an
11-cell grid --- 5 cue dimensions $\times$ 2 poles (functional/hedonic), plus one neutral cell ---
which the sweeps then cross with price multipliers and description
variants.

\subsection{Models and inference}
\label{app:models}

Local runs apply each model card's recommended chat template. We measure the first-token log-probability of the ``A''/``B'' answer tokens. GPT~5.4 and Gemini 3.5 Flash expose no token logits, so they enter only choice rate analyses.

\subsection{LLM-judge scores}
\label{app:quality}

\paragraph{Quality.} 
We use GPT 5.4 as a numeric judge to produce quality scores. Each option is rated five times on a $[0, 100]$
scale and the ratings are averaged. The judge is instructed to rate the product description, and prices are not provided. \Cref{tab:quality_tiers} (Panel~A) confirms that generation produces a
smoothly increasing quality ladder; Panel~B shows the mean rating across
the five quality-sweep perturbations ($y_{\mathrm{index}} \in
\{-2,\dots,+2\}$), each of which re-generates the option description with
different attributes while holding price fixed.

\begin{figure*}[h!]
    \centering \includegraphics[width=0.7\linewidth]{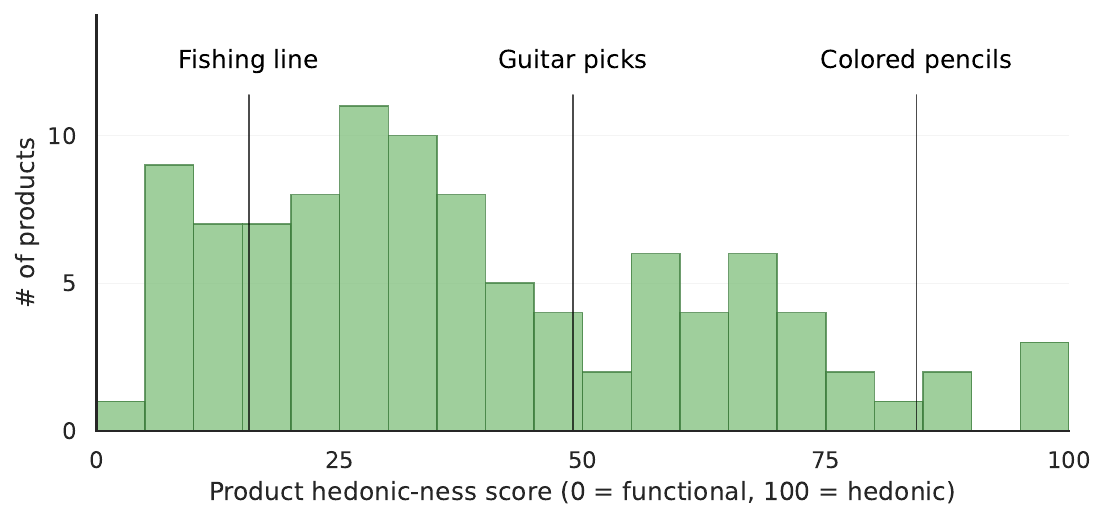}
    \caption{\textbf{Distribution of hedonic-ness scores across the 100
    products.} Scores from the GPT~5.4 judge
    (\Cref{app:quality}); 0 = purely functional, 100 = purely hedonic. The
    distribution is right-skewed (median 33), with labeled products
    marking representative points.}
    \label{fig:appendix_hedonic_scores}
\end{figure*}

\paragraph{Hedonic-ness.} The product-level hedonic-ness scores of
\Cref{sec:categorization} also come from a GPT 5.4-based judge \citep{asirvatham_gpt_2026}. Scores span the full range but are
right-skewed (median 33) (\Cref{fig:appendix_hedonic_scores}).

\section{Within-Model Standardization}
\label{app:standardize}

Models differ by an order of magnitude in the variance of
$\ell_i = \mathrm{logit}\,P(\text{premium})$ on neutral trials. For instance, Gemma 27B's
neutral standard deviation (SD) is roughly 5$\times$ Qwen 4B's. To compare shapes of cue
effects across models, we rescale every model's $\ell_i$ by its own
neutral-trial SD,
\begin{align*}
\tilde{\ell}_{i,m} = \frac{\ell_{i,m}}{\widehat{\sigma}_m^{(0)}}, \;  \widehat{\sigma}_m^{(0)} = \mathrm{SD}\!\big(\{\ell_i : i \in \mathrm{neutral}_m\}\big)
\end{align*}
so $\tilde{\ell}$ is in neutral-SD units within model $m$. All
``$\Delta$ std.\ logit'' axes in the paper (\Cref{fig:choice_rate} right,
\Cref{fig:frequency} right, \Cref{fig:scaling} right) use this
standardization, and pooled lines average $\tilde{\ell}$ across models with
equal weight.

\section{Reproducibility Statement}
\label{app:repro}

\paragraph{Compute.}
Local inference runs on NVIDIA H100 and H200 GPUs.
The benchmark requires approximately 1000 H100-hours for the Gemma 3 and Qwen 3 families. API costs for GPT 5.4, Gemini 3.5 Flash, GLM 5.2, DeepSeek v4 Pro, and Qwen 3.7 Plus totaled roughly \$500.

\paragraph{Software.}
Local inference uses \texttt{torch} 2.9.1, \texttt{transformers} 4.57.1,
and \texttt{sglang} 0.5.9; analysis uses \texttt{numpy}, \texttt{pandas},
\texttt{statsmodels} 0.14, \texttt{scipy}, \texttt{matplotlib}, and
\texttt{seaborn}. The full pinned environment is released with the code.

\paragraph{Licenses and intended use.}
Gemma 3 weights are used under the Gemma Terms of Use and Qwen 3 weights
under Apache 2.0; GLM 5.2, DeepSeek v4 Pro, and Qwen 3.7 Plus are accessed
through the Fireworks API, and GPT~5.4 and Gemini 3.5 Flash through their
providers' APIs, each under the respective terms of service. All use is
non-commercial research. The questions and model
responses will be released under CC~BY~4.0. The dataset is intended for research on LLM behavior, not for training of commercial systems.

\paragraph{AI assistance.}
Claude Sonnet~4.6 generated the product descriptions, prices, and
per-product cue instantiations (\Cref{app:pairs,app:cues}); the generation
prompts and parameters are released with the dataset. GPT~5.4 served as a
numeric judge for quality and hedonic-ness scores (\Cref{app:quality}).
The authors used Claude and ChatGPT for coding assistance and
for editing portions of the manuscript. All scientific
decisions were made by the authors.

\paragraph{Risks and privacy.}
One risk is that the cue effects we document could be read as a recipe
for steering LLM recommendations by adding framing effects that are invisible to users. That said, cues are borrowed from existing literature, and releasing the dataset adds little novel capability: all stimuli
are synthetic, brand-free, and contain no demographic, geographic, or
identity-related framing, and the manipulations are drawn from a
longstanding consumer choice literature. We manually audited all 100
products and a random sample of cue texts; no PII or offensive material
is present, and model responses contain no free-form text beyond A/B
choices and log-probabilities.

% =============================================================================
% Appendix figures
% =============================================================================

% --- begin inlined: tab_pair_examples.tex ---

\begin{figure*}[h!]
    \centering \includegraphics[width=0.8\linewidth]{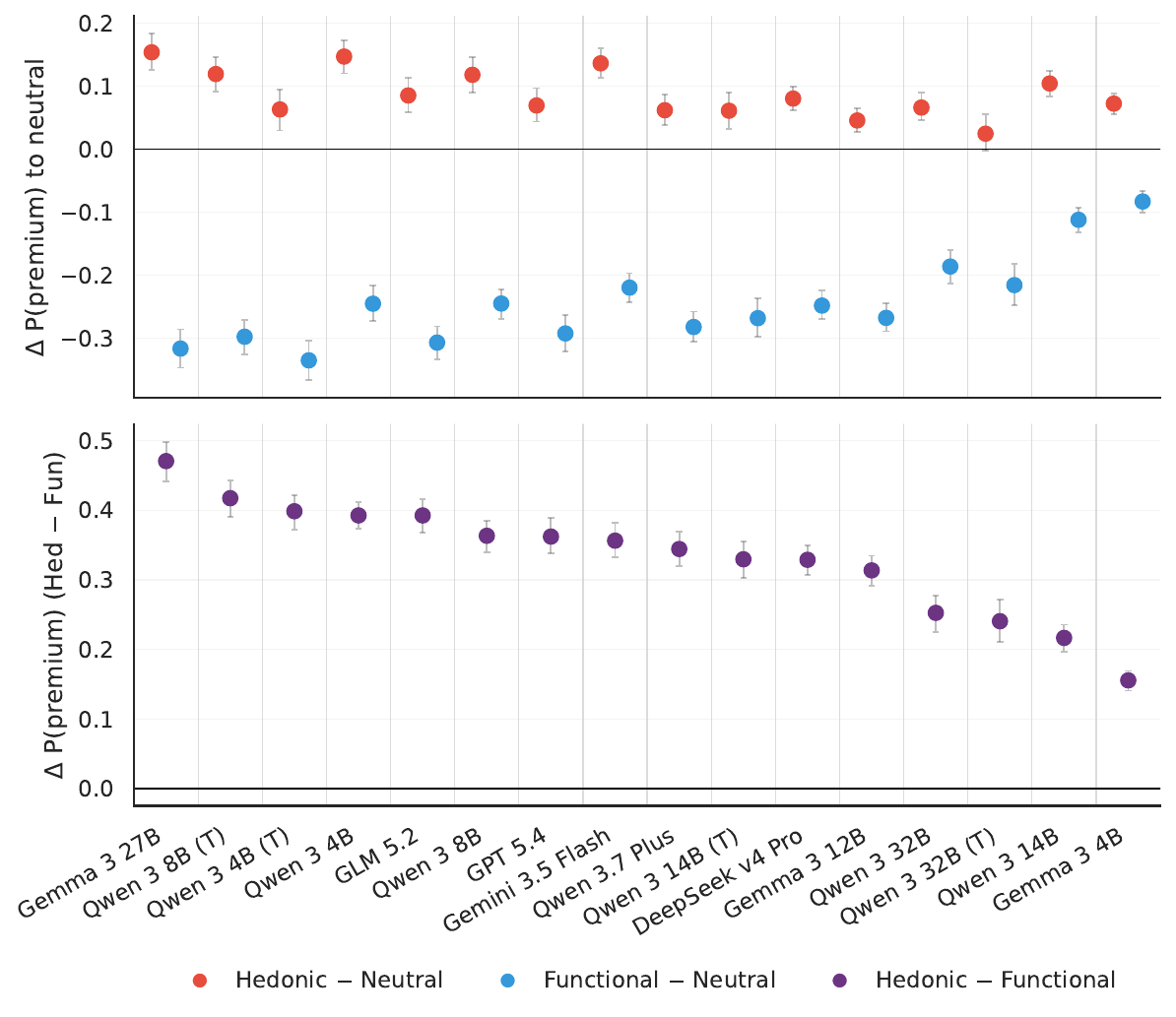}
    \caption{\textbf{Premium choice rates by model and cue condition on the baseline
    dataset}, for all sixteen configurations (twelve models plus the four
    thinking modes; (T) denotes thinking mode). (\textbf{Top}) Difference in premium choice rate between hedonic (functional) cues and the neutral condition. (\textbf{Bottom}) Difference in premium choice rate between hedonic and functional cues, averaged across cue types. 
    All bars are 95\% CIs.}
    \label{fig:appendix_hed_fun}
\end{figure*}

\begin{figure*}[h!]
    \centering \includegraphics[width=\linewidth]{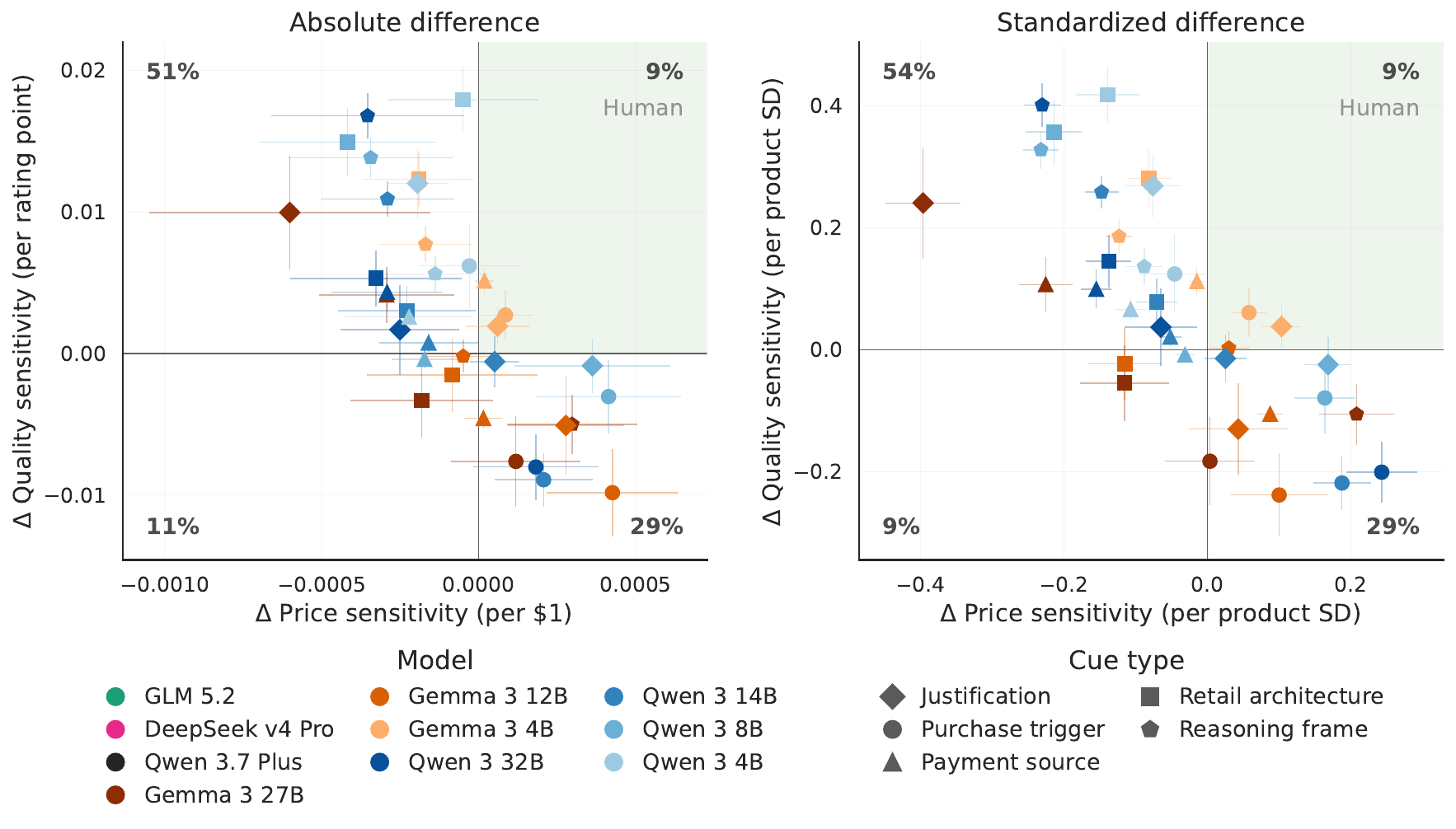}
    \caption{\textbf{Cue effects on feature sensitivity under alternate
    regression specifications.} The same quantity as \Cref{fig:choice_rate}
    (right), re-estimated with differences in feature values instead of log
    ratios: absolute differences (left; price sensitivity per dollar,
    quality sensitivity per rating point) and per-product-SD standardized
    differences (right). Percentages report the share of model-cue
    combinations in each quadrant.}
    \label{fig:appendix_regression_robustness}
\end{figure*}

\begin{figure*}[h!]
    \centering \includegraphics[width=0.7\linewidth]{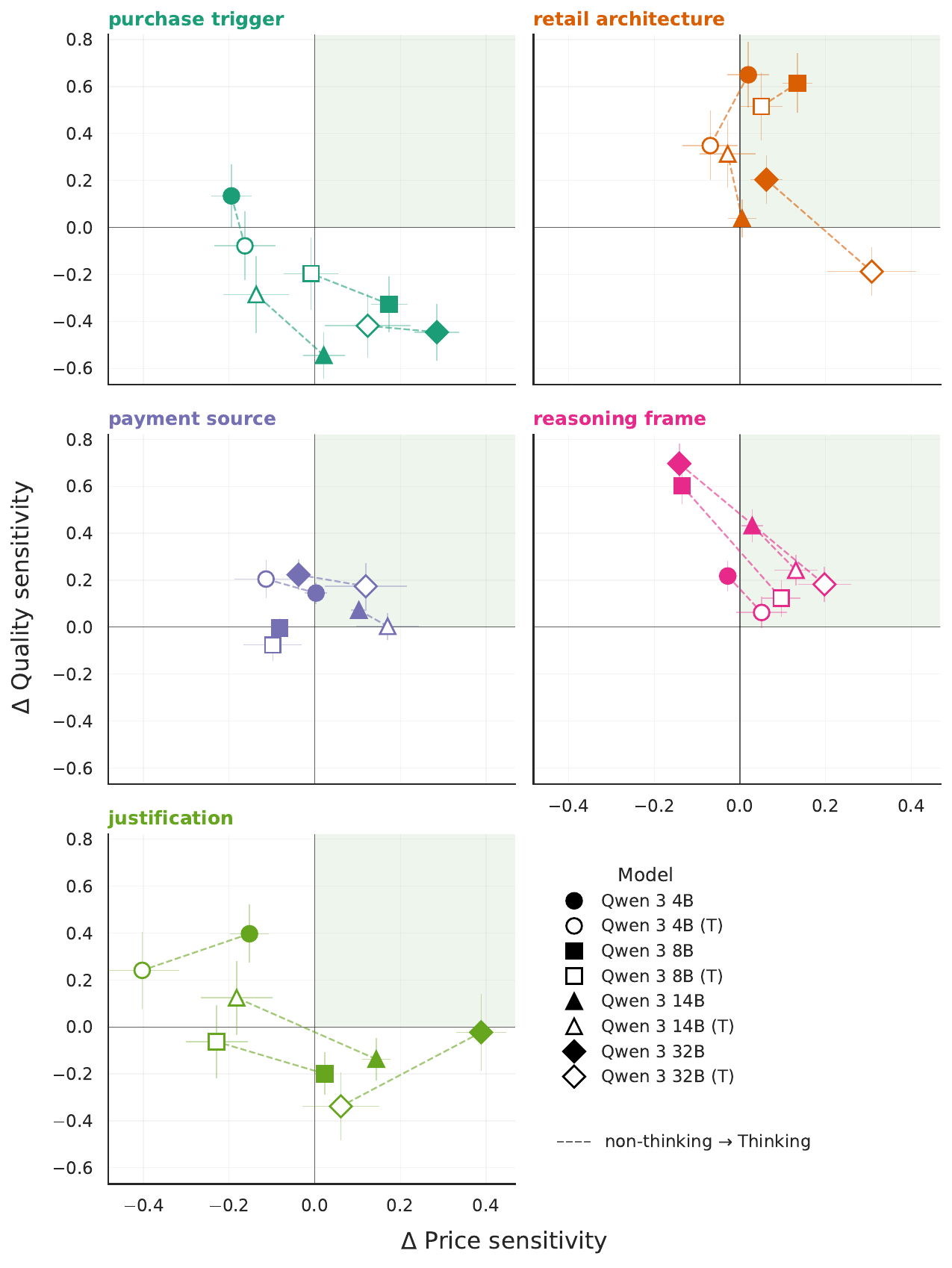}
    \caption{\textbf{Effect of thinking mode on price and quality
    sensitivity, by cue type.} Difference in standardized price (x-axis) and quality (y-axis) sensitivity between hedonic and functional cues for Qwen models in thinking/non-thinking modes. Shaded region denotes human-consistent behavior. \Cref{fig:scaling} (right) shows the justification and reasoning frame dimensions. All bars are 95\% CIs.}
    \label{fig:appendix_qwen_impact}
\end{figure*}

\clearpage

\begin{table*}[h!]
\centering
\small
\setlength{\tabcolsep}{8pt}
\renewcommand{\arraystretch}{1.25}
\begin{tabularx}{\textwidth}{@{}l X r@{}}
\toprule
\multicolumn{3}{@{}l}{\textbf{Fishing reel} \quad \textit{outdoor \& sporting goods} \quad\textbar\quad price ratio 2.33} \\
\midrule
\emph{Premium} & \textbf{Aluminum spinning reel with sealed drag.} Aluminum alloy body and rotor, 6 ball bearings, machined aluminum spool with sealed multi-disc drag, suited for light saltwater fishing. & \$69.99 \\
\emph{Standard}  & \textbf{Graphite spinning reel, entry level.} Graphite composite body and rotor, 3 ball bearings, aluminum spool with basic anti-reverse mechanism, suited for freshwater fishing. & \$29.99 \\
\midrule
\multicolumn{3}{@{}l}{\textbf{Electric toothbrush} \quad \textit{personal care \& grooming} \quad\textbar\quad price ratio 3.13} \\
\midrule
\emph{Premium} & \textbf{Rechargeable oscillating toothbrush.} Single oscillating speed, USB charging cable, plastic handle, one replacement brush head and two-minute timer. & \$24.99 \\
\emph{Standard}  & \textbf{Basic battery-powered toothbrush.} Single-speed vibrating motor, AA-battery powered, fixed plastic handle, one nylon bristle brush head included. & \$7.99 \\
\bottomrule
\end{tabularx}
\caption{\textbf{Two example product pairs.} Each row shows the option name (bolded) and description generated by Claude Sonnet 4.6, along with its price. Descriptions are length-matched and brand-free, and option order (A=premium vs.\ A=standard) is counterbalanced across trials.}
\label{tab:pair_examples}
\end{table*}

\begin{table*}[h!]
\centering
\small
\setlength{\tabcolsep}{8pt}
\renewcommand{\arraystretch}{1.3}
\begin{tabularx}{\textwidth}{@{}l l X@{}}
\toprule
\textbf{Dimension} & \textbf{Pole} & \textbf{Cue} \\
\midrule
\multirow{2}{*}{\textbf{Purchase trigger}}
  & \emph{Hed.}  & I'm shopping for a chef's knife because I've been curious about the category and I'm considering an upgrade to my everyday cooking setup. \\
  & \emph{Func.} & I'm shopping for a chef's knife because my current one has worn down and I want a solid replacement without overpaying. \\
\midrule
\multirow{2}{*}{\textbf{Retail architecture}}
  & \emph{Hed.}  & I'm browsing for a chef's knife on a curated kitchenware shop page where staff picks include quality notes on blade feel and durability. \\
  & \emph{Func.} & I'm browsing for a chef's knife on a marketplace comparison page sorted by price, with filters, unit costs, and price-match language displayed. \\
\midrule
\multirow{2}{*}{\textbf{Payment source}}
  & \emph{Hed.}  & I'm looking for a chef's knife and will be paying with a gift card I received as a birthday present. Any unspent amount remains mine. \\
  & \emph{Func.} & I'm looking for a chef's knife and will be paying with money from my regular checking account. Any unspent amount remains mine. \\
\midrule
\multirow{2}{*}{\textbf{Justification}}
  & \emph{Hed.}  & I'm looking for a chef's knife. I recently completed a demanding project and earned a small reward, so this feels like a reasonable treat. \\
  & \emph{Func.} & I'm looking for a chef's knife. This is an ordinary purchase with no special reason to treat myself---I'm just making a practical choice. \\
\midrule
\multirow{2}{*}{\textbf{Reasoning frame}}
  & \emph{Hed.}  & I'm looking for a chef's knife. I'm focused on why this product would improve my use experience and the qualities I'd notice while using it. \\
  & \emph{Func.} & I'm looking for a chef's knife. I'm focused on how I'll complete the purchase and the concrete tradeoffs to compare. \\
\bottomrule
\end{tabularx}
\caption{\textbf{Cue templates instantiated for one product} (chef's knife). Each of the five dimensions has a paired hedonic/functional cue. Cues for the remaining 99 products are generated analogously via Claude Sonnet 4.6. Appendix \ref{app:prompt} provides the full prompt template for choice questions.}
\label{tab:cue_templates}
\end{table*}

\clearpage
\begin{table*}[h!]
\centering
\footnotesize
\setlength{\tabcolsep}{4pt}
\begin{tabular}{lrrrr}
\toprule
Family & Trials & H. & F. & N. \\
\midrule
Baseline      &   8{,}800 &  4{,}000 &  4{,}000 &    800 \\
Price sweep   &  88{,}000 & 40{,}000 & 40{,}000 &  8{,}000 \\
Quality sweep &  44{,}000 & 20{,}000 & 20{,}000 &  4{,}000 \\
% ICL priming   &  35{,}200 &        0 &        0 & 35{,}200 \\
\midrule
ICL mix       &  36{,}000 &        0 &        0 & 36{,}000 \\
\midrule
Total              & 176{,}800 & & & \\
\bottomrule
\end{tabular}
\caption{\textbf{Trial counts per model.} ``H.''/``F.''/``N.'' are hedonic,
functional, and neutral conditions. The same 800 option pairs are reused
across all four experiment families, but not every model is evaluated in each experiment.}
\label{tab:trial_counts}
\end{table*}

\begin{table*}[h!]
\centering
\small
\begin{tabular}{cllrr}
\multicolumn{5}{l}{\textbf{Panel A: baseline tier ratings ($y_{\mathrm{index}}=0$)}} \\
\toprule
Tier & Label & Mean & SD & $n$ \\
\midrule
1 & budget & 30.4 & 9.2 & 100 \\
2 & economy & 54.6 & 11.8 & 100 \\
3 & midrange & 72.9 & 8.7 & 100 \\
4 & premium & 84.8 & 4.3 & 100 \\
5 & luxury & 89.8 & 3.7 & 100 \\
\bottomrule
\end{tabular}

\vspace{6pt}
\begin{tabular}{lrrrrr}
\multicolumn{6}{c}{\textbf{Panel B: quality-sweep mean rating}} \\
\toprule
Tier & $-2$ & $-1$ & $0$ & $+1$ & $+2$ \\
\midrule
1 (budget) & 19.4 & 28.2 & 30.4 & 41.4 & 53.1 \\
2 (economy) & 31.7 & 46.6 & 54.6 & 61.7 & 67.5 \\
3 (midrange) & 49.3 & 65.9 & 72.9 & 76.6 & 81.0 \\
4 (premium) & 67.5 & 80.3 & 84.8 & 86.5 & 88.6 \\
5 (luxury) & 81.7 & 87.5 & 89.8 & 90.8 & 92.4 \\
\bottomrule
\end{tabular}
\caption{\textbf{GPT~5.4 quality ratings.} Each option is
rated five times independently from 0 to 100 with a rubric that ignores price and brand
and scores only descriptive attributes (\Cref{app:quality}). Panel A: mean
rating per baseline tier across the 100 products
($y_{\mathrm{index}}=0$). Panel B: mean rating across the five
quality-sweep perturbations.}
\label{tab:quality_tiers}
\end{table*}

\begin{table*}[h!]
\centering
\small
\begin{tabular}{lllll}
\toprule
Model & Access & Backend & Decoding & Choice DV \\
\midrule
Gemma 3 4B/12B/27B-IT & Open & Local (\texttt{transformers}) & greedy & first-token A/B logit \\
Qwen 3 4B/8B/14B/32B & Open & Local (SGLang) & greedy & first-token A/B logprob \\
Qwen 3 4B--32B (Thinking) & Open & Local (SGLang) & T=0.6, top-p 0.95, 4096-tok CoT & A/B logprob after CoT \\
GLM 5.2 & API & Fireworks (reasoning off) & T=0 & A/B logprob (quantized) \\
DeepSeek v4 Pro & API & Fireworks (reasoning off) & T=0 & A/B logprob (quantized) \\
Qwen 3.7 Plus & API & Fireworks (reasoning off) & T=0 & A/B logprob (quantized) \\
Gemini 3.5 Flash & API & Google AI Studio & default & choice only \\
GPT 5.4 & API & OpenAI Batch & reasoning = medium & choice only \\
\bottomrule
\end{tabular}
\caption{\textbf{Models evaluated.}}
\label{tab:models}
\end{table*}

\end{document}